\documentclass[10pt,twocolumn,letterpaper]{article}

\usepackage{wacv}
\usepackage{times}
\usepackage{epsfig}
\usepackage{graphicx}
\usepackage{amsmath}
\usepackage{amssymb}
\usepackage{booktabs}   
\usepackage{lipsum}
\usepackage{multirow}
\usepackage{array}
\usepackage{caption}
\usepackage{subcaption}
\usepackage{enumitem}

\def\wacvPaperID{1628} 

\wacvalgorithmstrack   

\wacvfinalcopy 

\ifwacvfinal
\usepackage[breaklinks=true,bookmarks=false]{hyperref}
\else
\usepackage[pagebackref=true,breaklinks=true,colorlinks,bookmarks=false]{hyperref}
\fi
\usepackage{authblk}

\begin{document}



\title{One-Shot Doc Snippet Detection: Powering Search in Document Beyond Text}


\author[]{Abhinav Java$^*$}
\author[]{Shripad Deshmukh$^{*}$}
\author[]{ Milan Aggarwal}
\author[]{Surgan Jandial}
\author[]{Mausoom Sarkar}
\author[]{Balaji Krishnamurthy}

\affil[]{Adobe, Media and Data Science Research Labs, Noida, India -- 201304 \authorcr
  \{\tt ajava, shdeshmu, milaggar, jandial, msarkar, kbalaji\}@adobe.com}

\date{\today}

\maketitle
\def\thefootnote{*}\footnotetext{These authors contributed equally to this work}\def\thefootnote{\arabic{footnote}}

\thispagestyle{empty}

\begin{abstract}

Active consumption of digital documents has yielded scope for research in various applications, including search. Traditionally, searching within a document has been cast as a text matching problem ignoring the rich layout and visual cues commonly present in structured documents, forms, etc. To that end, we ask a mostly unexplored question: ``Can we search for other \emph{similar} snippets present in a target document page given a single query instance of a document snippet?". We propose \emph{MONOMER} to solve this as a one-shot snippet detection task. \emph{MONOMER} fuses context from visual, textual, and spatial modalities of snippets and documents to find query snippet in target documents. We conduct extensive ablations and experiments showing \emph{MONOMER} outperforms several baselines from one-shot object detection (BHRL), template matching, and document understanding (LayoutLMv3). Due to the scarcity of relevant data for the task at hand, we train \emph{MONOMER} on programmatically generated data having many visually similar query snippets and target document pairs from two datasets - Flamingo Forms and PubLayNet. We also do a human study to validate the generated data.  

\end{abstract}

\section{Introduction}

\begin{figure}[h]
    \centering
    \includegraphics[width=0.9\linewidth]{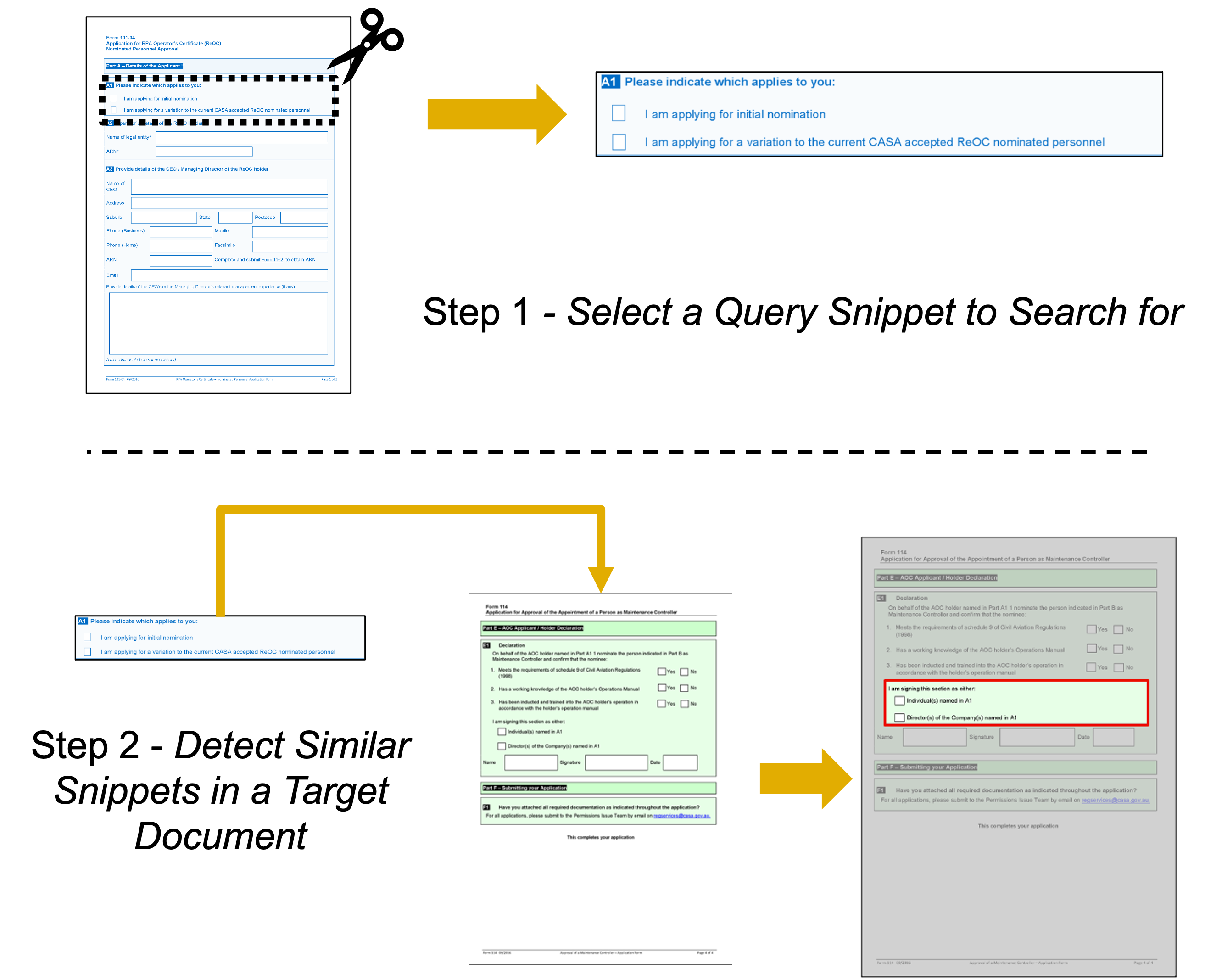}
    \caption{A new paradigm for search in documents through one-shot snippet detection.}
    \label{fig:task_overview}
\end{figure}



Documents have been the primary medium for storing and communicating information in academia, public offices, private businesses, print media, etc \cite{macdonald2001using}. With world transitioning to a digital-first ecosystem, expedited by the challenges posed by the ongoing pandemic \cite{oecd2020digital}, the trends in document usages are shifting from passive modes like reading/sharing documents to more active modes such as authoring a document, editing the styles, customising figures and tables, etc \cite{distefano_adobe_2020, document_cloud_how_2022, modi_adobe_2020}. 
However, search functionality within documents is mostly limited to locating regions in a page containing text that matches a given textual query \cite{Berry2007SurveyOT, 10.1145/2071389.2071390, tito2021icdar}. Confining search to textual modality restricts several use cases. For instance, consider scenarios where a form author wants to search for binary male/female options in personal detail sections to add one more option to collect the gender information, or a document editor wants to search for image-caption pairs to swap their ordering, or a situation wherein a student wants to search for peculiar information stored in special kind of tables. These scenarios emphasise the need for more advanced search capabilities based on document snippets.


Hence, a utility which allows users to select a rectangular snippet in a page, and find other similar snippets in a target document would be a stepping stone towards empowering this search experience. To achieve this, we model it as \textit{one-shot doc snippet detection} task i.e. detect regions in a target document page that are similar to a given snippet (as shown in Fig.~\ref{fig:task_overview}). 
Existing text-based search tools \cite{searching_pdf_2021} are incapable of detecting visually similar snippets due to their inherent design as they lack mechanism to incorporate visual and layout cues. On the other hand, document structure extraction methods \cite{aggarwal2020multi, lin2017feature, xu2020layoutlm} are trained to identify predefined generic class structures (such as a paragraph) in a document and hence, cannot be applied directly to detect arbitrary snippet patterns in one-shot setting. Furthermore, document image retrieval tasks such as logo-detection \cite{6831022}, signature field retrieval \cite{kiran2021offline} etc. are designed to extract task-specific entities like logos and sign-fields respectively. 


To attain the capability of ``search with snippets'' in documents, we formulate the problem as a one-shot snippet detection task and design a multi-modal method for the same. We propose a multi-modal framework -- \emph{\textbf{MONOMER}} (\textbf{M}ultimodal Cr\textbf{O}ss Atte\textbf{N}tion-based D\textbf{O}cu\textbf{M}ent Snipp\textbf{E}t Sea\textbf{R}ch) that fuses context from visual, textual and spatial modalities across the query snippet and target document through attention (Section~\ref{sec:model_details}). The fused representations are then processed through a feature pyramid network, followed by region proposal predictions, to identify the boundaries of regions in the target document that are similar to the query snippet. We compare our approach with the current state-of-the-art method in One-Shot Object Detection - BHRL \cite{yang2022balanced} and a task-specific extension of the best performing document analysis approach - LayoutLMv3 \cite{huang2022layoutlmv3} in Table~\ref{tab:comparison_results}. We show that MONOMER outperforms the above baselines (Section~\ref{sec:comp_baselines}), highlighting the effectiveness of our proposed framework. In Section~\ref{sec:ablation} we demonstrate the advantage of using all three modalities by performing extensive ablations with various modality combinations.

The scarcity of relevant data poses an additional challenge in tackling the task. Documents in the form of images, text, and layout\cite{bowen2009document} are widely available. However, annotated data of document snippets and their associated matching regions in other documents are hard to find. Adding to this, different modalities like visual, layout and textual imply similarity in a highly subjective manner. This makes obtaining large-scale human-annotated data for snippet search extremely challenging. To make the problem tractable, we design a programmatic way of obtaining similar query snippet and target document pairs by defining similarity based on alignment between the layout of their basic constituent structures. More specifically, we sort the constituent structures (such as Text, Tables, Fillable areas etc.) in a doc snippet according to natural reading order, followed by creating a layout string based on the sequential sorted ordering. Likewise, we obtain the layout string corresponding to each document in the corpus. The snippet is associated with documents whose layout string has at least one contiguous subsequence that aligns with the snippet's layout string. Therefore, we propose a layout-centric definition of similarity that enforces alignment between the layout of two snippets to be deemed similar, which provides an automated technique for emulating human labeling. We choose Flamingo forms \cite{sarkar2020document} and PubLayNet documents \cite{zhong2019publaynet} as the underlying corpora to create two similarity matching datasets. We discuss the data generation procedure in detail, followed by its validation through human study in Section~\ref{dataset}. 
To summarize, the contributions of this work can be enumerated as the following:

\begin{itemize}[noitemsep,topsep=0pt]
    \item We formulate the task of one-shot document snippet detection for advancing search in document domain beyond traditional text-based search.
    \item We define layout based document-snippet similarity that allows generation of similarity matching data at scale in a fully programmatic manner, the validity of which is supported by an extensive human study. We plan to release a part of the introduced datasets.
    \item We propose MONOMER, a multi-modal framework for snippet detection that performs better than one-shot object detection and multi-modal document analysis baselines. Further, MONOMER is able to perform well on layout patterns not seen during training.
\end{itemize}

\section{Related Work}

\subsection{Document Understanding}

Understanding documents requires comprehending the content present in document page i.e. images, text, and any other multi-modal data in conjunction with the layout, structures, placement of the content, blank spaces, etc. For understanding content, prior research works have designed tasks such as DocVQA \cite{tito2021icdar}, InfographicsVQA \cite{mathew2022infographicvqa} etc. while layout understanding has been formally studied through Document Layout Analysis \cite{aggarwal-etal-2020-form2seq, 10.1145/3355610}. Layout analysis has been formulated as an object detection task \cite{zaidi2022survey} to extract structures such as headings, tables, text blocks, etc. from a document image. Such approaches extensively use state-of-the-art object detection heads (for eg. YOLO \cite{redmon2016you}, Faster-RCNN \cite{ren2015faster} etc.) usually employed in the domain of natural images. Methods such as HighResNet \cite{sarkar2020document}, MFCN \cite{yang2017learning} etc. approach Layout Analysis as pixel-level segmentation of document image. Subsequently, several recent works like DocFormer \cite{appalaraju2021docformer},  LayoutLM \cite{xu2020layoutlm}, DiT \cite{li2022dit} proposed large-scale pre-training techniques to cater the document understanding task. The representations learned by these models have turned out to be very useful in many downstream tasks, both for content understanding and layout parsing. In this work, we leverage such representations to develop snippet based search tools for the documents.

\subsection{Template Matching}
Template Matching refers to the task of detecting and localizing a given query image in a target (usually larger) image. Seminal template matching literature leverages traditional computer vision techniques like Normalized Cross Correlation (NCC) \cite{yoo2009fast} and Sum of Squares Differences (SSD) \cite{7449303} for searching. Despite their widespread success, the aforementioned techniques have clear limitations in regard to matching templates which are complex transformations of the instance present in the target image. For instance, NCC/SSD might fail due to large variation in scale, occlusions etc. Consequently, feature matching-based techniques such as SIFT \cite{wu2013comparative} and SURF \cite{oyallon2015analysis} were proposed to allow matching local features between images to address scale-invariance. Typically, these methods find local key-points in images. Yet, several issues like image quality, lightning, real-time use severely limit the applicability of these approaches. The recent surge of Deep Learning allowed researchers to develop more sophisticated techniques like QATM \cite{cheng2019qatm}, DeepOneClass \cite{ruff2018deep} that perform Siamese matching between deep features of natural images for tasks like GPS localization. QATM \cite{cheng2019qatm} propose a learnable matching layer that enables better matching within natural images compared to standard Siamese matching. However, we note that matching templates within documents is a different (from natural images) and non-trivial task with additional nuances owing to the diverse and complicated arrangement of layout, visual structures and textual content contained in a document.

\subsection{One Shot Object Detection (OSOD)}
OSOD aims at detecting instances of novel classes (not seen during training) within a test image given a single example of the unseen/novel class. At a high level, most OSOD techniques perform alignment between deep features of query (example of novel class) and target image (test image where the novel class instance is present). Recently, methods such as COAE \cite{hsieh2019one}, AIT \cite{chen2021adaptive} etc. have shown that the learned attention-based correlation can outperform standard Siamese matching \cite{li2018high, melekhov2016siamese} since they capture multi-scale context better through global and local attention. Popular OSOD techniques \cite{lin2014microsoft} have been shown to perform well on natural images when class definitions are clearly specified. However, due to the complexity of document data and lack of well-defined yet exhaustive set of layout patterns, it is not possible to enumerate a finite set of classes. More recently, \cite{yang2022balanced} proposed a technique to learn a hierarchical relationship (BHRL) between object proposals within a target and the query. While BHRL shows impressive performance on natural images, it does not leverage multi-modal information that is critical for document snippet detection. Contrary to existing approaches, we leverage all possible co-relations between different modalities of query and target and show that we are able to achieve better overall performance on complex document data where existing methods typically fail.

\section{One-Shot Doc Snippet Detection}

\subsection{Problem Formulation}


We first give an overview of dataset creation and outline of the task formulation followed by their details.

\noindent\textbf{Dataset Creation.} Let $\mathcal{X}$ be the set of all document snippets. We define a similarity criterion  $g_{sim}: \mathcal{X}^2 \xrightarrow{} \mathcal{R}$ which takes two document snippets $A, B \in \mathcal{X}$, and outputs similarity score $s=g_{sim}(A, B)$. The $g_{sim}$ function can be defined according to human's notion of similarity, or as a fully programmatic similarity criterion. Query-target pairs $(Q, T)$ are mined from the document corpus using $g_{sim}$ such that $Q \in \mathcal{X}$ and target document $T$ contains a non-empty set of snippets  $\mathcal{S}_{qt}=\{S_i | S_i \in \mathcal{X}, g_{sim}(S_i, Q) > th_{sim}, i=1, 2, ..., n \}$; $th_{sim}$ being the threshold over similarity score. $(Q,T)$ pairs are collected to create dataset $\mathcal{D}$.

\noindent\textbf{Task Definition.} Given a dataset $\mathcal{D}$ of query-target pairs which are generated using an oracle $g_{sim}$ (not accessible afterwards), our task is to find $\mathcal{S}_{qt}$ for each pair $(Q,T) \in \mathcal{D}$. Let $f_{\theta}$ be a model with parameters $\theta$ which predicts similar snippets $\hat{S}_{qt}$ for given $(Q,T)$ pair and let loss $L$ be the measure of error between $S_{qt}$ and $\hat{S}_{qt}$. Then the doc snippet detection task becomes that of minimizing $L$ as follows 
\[
    \min_{\theta}\sum_{\forall (Q, T) \in \mathcal{D}} L(\mathcal{S}_{qt}, \hat{\mathcal{S}}_{qt})
\]

\subsection{Snippet-Document Dataset}
\label{dataset}


In this section, we discuss the details of how we define similarity in the context of documents using the layout of different snippets and documents to generate $\mathcal{D}$ followed by a human study to validate the quality of generated data.



\subsubsection{Dataset Generation}
\label{dataset_generation}

\begin{figure}[t]
    \centering
    \includegraphics[width=\linewidth]{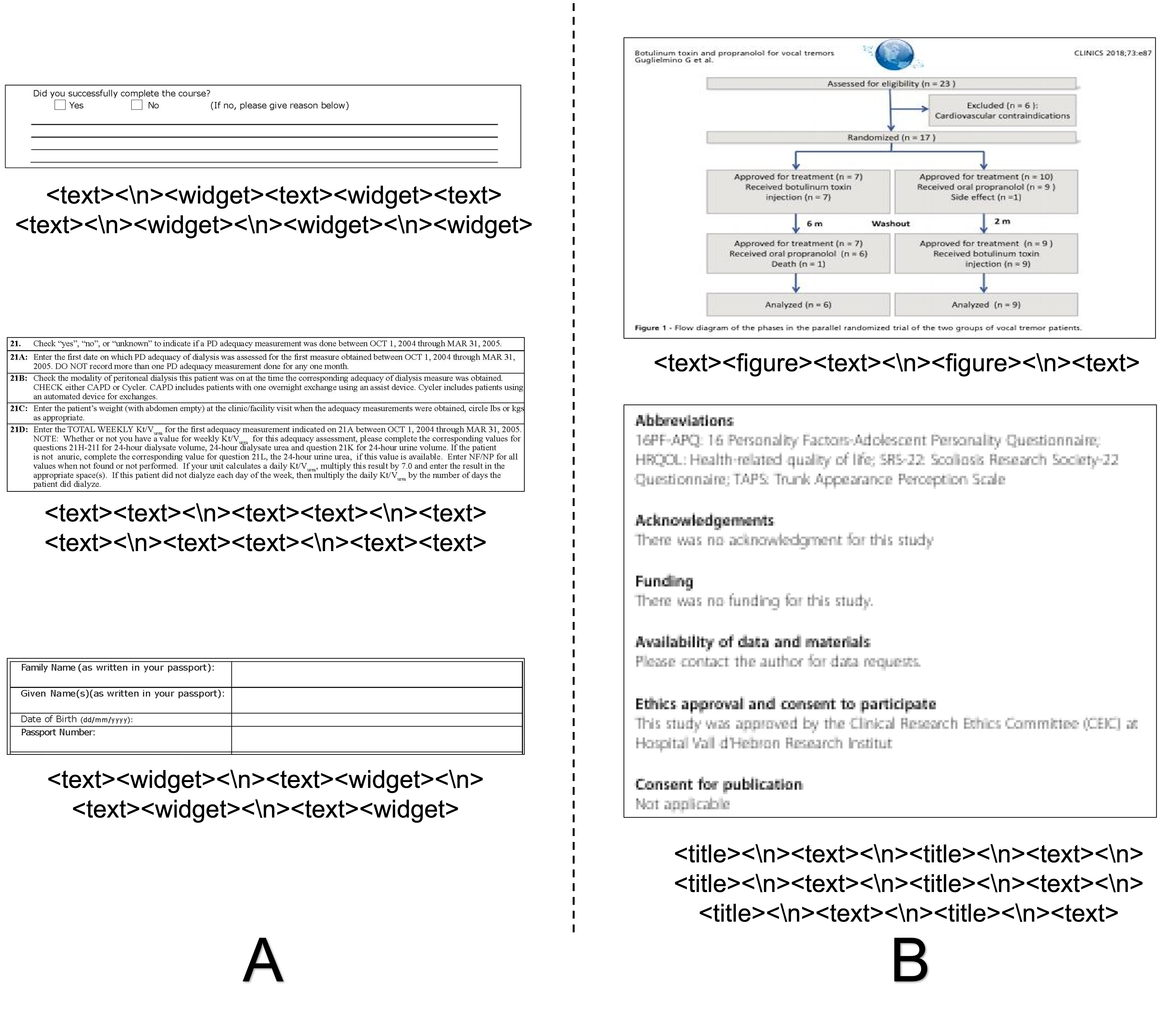}
    \caption{Snippets and the corresponding layout strings for A) Flamingo Forms,  B) PubLayNet Documents.}
    \label{fig:lstr_flm_plus_pub}
\end{figure}

\begin{figure*}
\begin{center}
\includegraphics[width=0.9\linewidth]{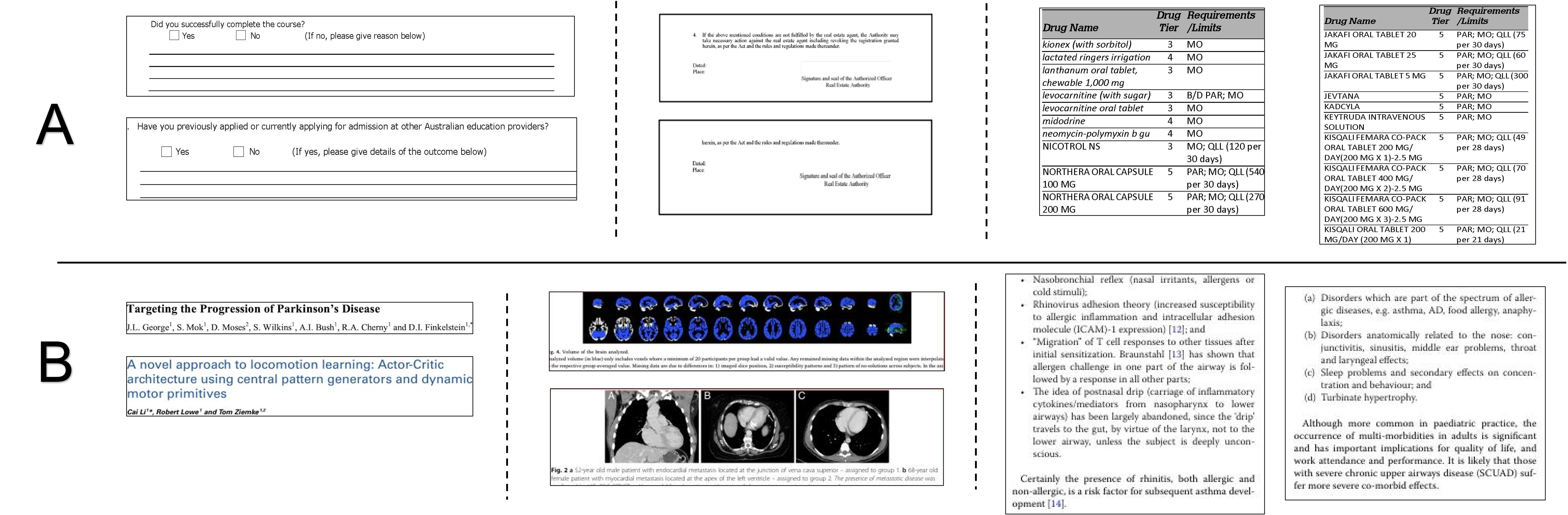}
\end{center}
\caption{Similar Snippets extracted programmatically for - A) Flamingo Forms, and B) PubLayNet Documents.}
\label{fig:similar_snippets}
\end{figure*}

Since document similarity depends on various factors and is highly prone to subjectivity, obtaining significant number of $(Q, T)$ pairs through human annotation becomes quite challenging. To that end, we decide to define $g_{sim}$ criterion programmatically as the following:

\begin{equation}
    g_{sim}(A, B) = 1 - \frac{d(lstr_{a}, lstr_{b})}{length(lstr_{a})}
    \label{eqn:data_formula}
\end{equation}
where $lstr_a$ and $lstr_b$ denote the layout strings of snippets A and B respectively and $d$ denotes the edit distance \cite{10.5555/1822502}. To obtain the layout string of a snippet or full document page, we sort their constituent structures (such as Text, Tables, Fillable areas etc.)\footnote{The bounds for basic elements are either present in the dataset or can be extracted using auto-tagging capabilities of
PDF tools.} according to natural reading order (top-bottom and left-right). We associate a symbol with each constituent element type such that the sequence of element symbols obtained according to the sorted ordering yields the layout string. Fig. \ref{fig:lstr_flm_plus_pub} shows examples of snippets and their corresponding layout strings. Given a snippet $\mathcal{Q}$ extracted randomly from some document, we provide its layout string as argument $lstr_a$ in Eq.~\ref{eqn:data_formula}. To identify if some other document in the corpus contains a similar region, we consider all possible contiguous subsequences of its layout string as candidates and provide the subsequence as input $lstr_b$ in Eq.~\ref{eqn:data_formula}. We filter candidates which qualifies the similarity threshold $th_{sim}$ of 0.92 (determined based on observation) to generate query-target pairs. ~Fig. \ref{fig:similar_snippets} illustrates the similar snippets identified using proposed $g_{sim}$.

We derive two similarity search datasets from two corpora -- Flamingo forms dataset\footnote{We contacted the authors to share their full dataset} \cite{sarkar2020document} and PubLayNet document dataset\footnote{\url{https://developer.ibm.com/exchanges/data/all/publaynet}} \cite{zhong2019publaynet}. The rationale behind choosing them is, a) forms data contain diverse layout structures with various hierarchical levels \cite{aggarwal2020multi, sarkar2020document}, and b) PubLayNet is a commonly used large scale documents dataset for document analysis. For Flamingo dataset, we use widgets (fillable area) and text blocks to create layout strings and for PubLayNet we consider text blocks, figures, lists, tables and titles as the layout symbols. Table \ref{tab:dataset_details} summarises number of samples obtained.

\begin{table}[]
\resizebox{235pt}{!}{%
\begin{tabular}{|c|cc|cc|}
\hline
\multirow{2}{*}{\textbf{Dataset}} & \multicolumn{2}{c|}{\textit{No. of (Q, T) pairs}} & \multicolumn{2}{c|}{\textit{Unique Layout Strings}} \\ \cline{2-5} 
 & \multicolumn{1}{c|}{Train} & Test & \multicolumn{1}{c|}{Train} & Test \\ \hline
\textit{Flamingo} & \multicolumn{1}{c|}{102065} & 24576 & \multicolumn{1}{c|}{6365} & 1911 \\ \hline
\textit{PubLayNet} & \multicolumn{1}{c|}{204256} & 15734 & \multicolumn{1}{c|}{35} & 23 \\ \hline
\end{tabular}}
\caption{Summary of snippet-document pairs datasets. Less number of unique layout strings in PubLayNet indicates limited combinations in which structures are organised.}
\label{tab:dataset_details}
\end{table}



\begin{figure*}[t]
    \centering
    \includegraphics[scale=0.95]{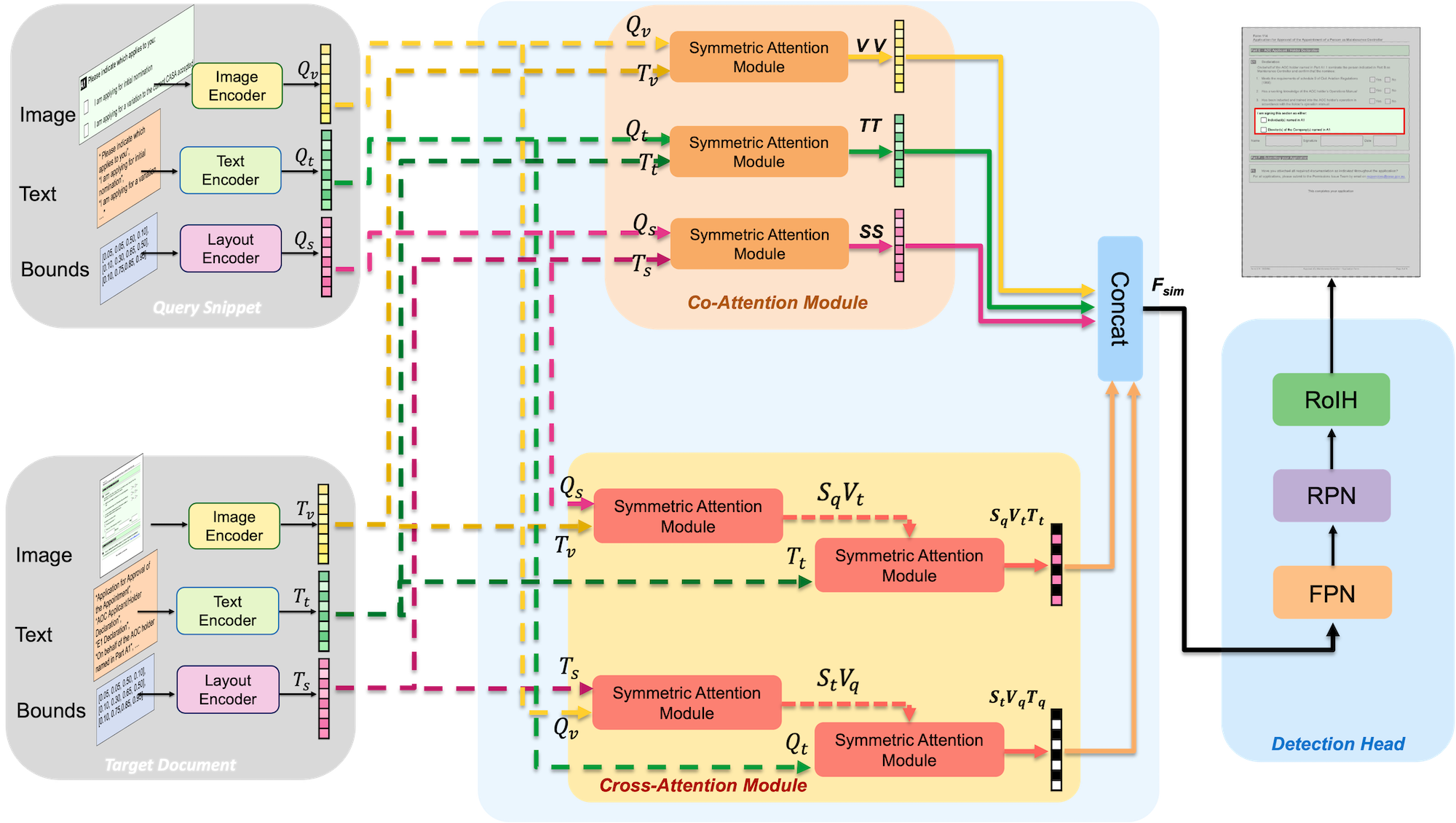}
    \caption{Architecture of the proposed MONOMER approach.}
    \label{fig:full_architecture}
\end{figure*}

\subsubsection{Human Study on Generated Data}
\label{human_study}

To validate the quality of the generated data, i.e. to evaluate how well the programmatically generated query-target pairs align with human notions of similarity, we conduct a human study involving 12 evaluators\footnote{Evaluators were remunerated appropriately for the evaluation task.}. We evaluate a total of 160 snippet-target document pairs sampled randomly from our dataset generated using Forms such that these samples are divided into 4 batches of 40 samples each. All samples in a batch are then evaluated by 3 evaluators based on the following criteria - given regions in a target document, count the number of regions 1) that are highlighted as similar and are actually similar, 2) that are similar but not highlighted, 3) which are highlighted as similar but are not exactly the same as the snippet. The evaluators are also asked if the layout pattern of the snippet is hard. Based on above, we estimate batch-wise metrics such as precision, recall etc. and report the average across the batches\footnote{Please refer to supplementary for batch-wise evaluation report}. It is found that precision is $87.96\%$ i.e. in $\sim88\%$ cases, target document snippet highlighted as similar to query snippet by our method is actually similar; recall is found to be $81.07\%$ which indicates that $\sim81\%$ of actually similar regions are highlighted by our method. Further, it is found that $87.48\%$ of similar matches are the ones where target document region is not exactly the same as the snippet showing that our technique mostly identifies similar but not trivially exact matches. Finally, it is observed that $48.12\%$ of snippets comprise of layout pattern which is complicated and hard to search.



\section{MONOMER}
\label{sec:model_details}


As information in a document is mainly present in the form of images, text and layout, paradigms that leverage all the modalities simultaneously have turned out to be successful in the past. For instance, document analysis methods such as DocFormer \cite{appalaraju2021docformer}, SelfDoc \cite{li2021selfdoc}, LayoutLMv3 \cite{huang2022layoutlmv3} etc. have developed pre-trained multi-modal architectures achieving great results on a wide variety of tasks like layout extraction, text recognition, document image classification, table detection etc. Motivated by this, we design our framework with the aim of enabling it to pool context from various document modalities to perform one-shot snippet detection task.


A possible way to leverage multi-modal context is to directly use one of the aforementioned pre-trained models to obtain multi-modal embeddings for both query snippet and target document separately. However, doing so restricts interconnecting individual modalities between query snippet and target page. We substantiate this intuition empirically in Table~\ref{tab:comparison_results} by comparing our method against fine-tuning pre-trained multi-modal baselines for doc snippet detection. Hence, we embed each modality for both snippet and target document separately using image, text and layout encoders and propose to process them further through attention based interconnections between them. We now discuss our architecture (Fig.\ref{fig:full_architecture}) in detail.


\noindent\textbf{Feature Extraction.}
To encode the snippet and target document image, we use DiT \cite{li2022dit}, a visual-only document analysis model. The text present in query and target is encoded using BERT \cite{devlin2018bert} based text encoder. Additionally, we generate features for bounding box information of constituent elements of snippet (arranged in a sequence according to reading order). Specifically, we use a transformer \cite{vaswani2017attention} based encoder-only module which tokenizes box co-ordinates and embeds the sequence of bounds. Previously, methods like \cite{reed2022generalist, xu2020layoutlm} have used such an approach to process various types of sequential data. Consequently, we generate 6 types of embeddings in total (3 modalities of the query and 3 modalities of the target). The visual, textual and spatial embeddings for query snippet are denoted by $Q_v$, $Q_t$ and $Q_s$ respectively and likewise, for the target document we have $T_v$, $T_t$ and $T_s$.



\begin{figure*}
    \centering
    \includegraphics[height=4cm]{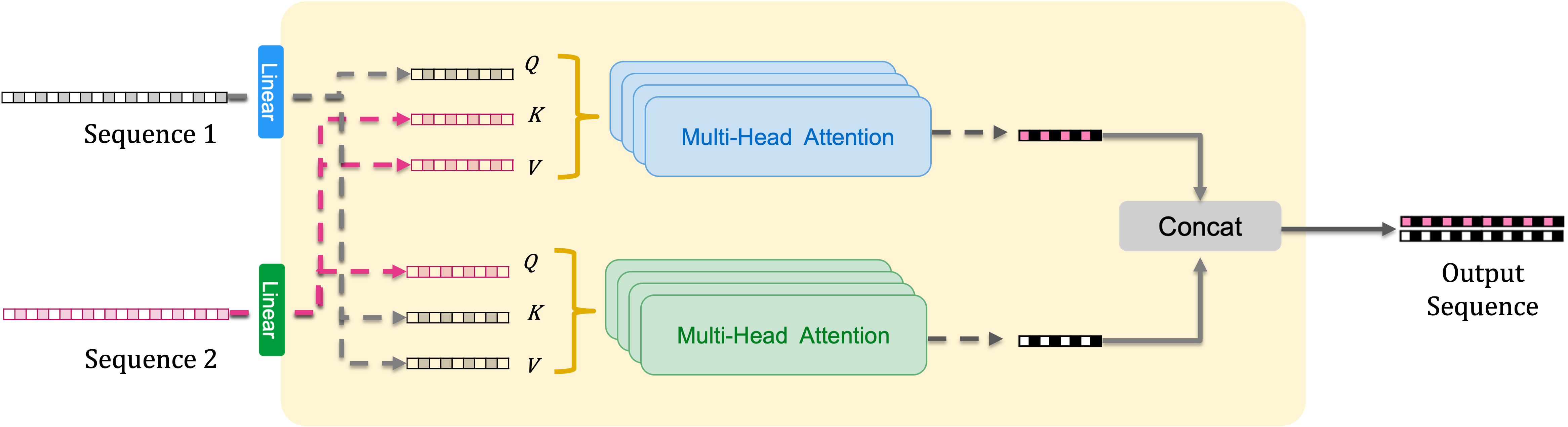}
    \caption{Architecture of Symmetric Attention Module.}
    \label{fig:symmetric_attention}
\end{figure*}



Parenthetically, the visual and spatial embedding of a document are highly interconnected as bounding boxes of documental entities determine the visual outlook of a page. Also, the visual embeddings from a vision-only model like DiT have demonstrated ability to detect and recognize text in downstream tasks \cite{li2022dit}, implying that the visual features also contain information about the text content of the document. Building on this intuition, we combine the extracted features accordingly.

\noindent\textbf{Query-Target Feature Fusion}
Features $Q_v$, $Q_t$, $Q_s$, $T_v$, $T_t$, and $T_s$ are in the form of token sequences outputted by corresponding transformer based encoders. We strategically apply \emph{symmetric attention} \cite{chen2021adaptive} between these token sequences. A symmetric attention of two sequences involves i) computing multi-head attention \cite{vaswani2017attention} of first sequence as query with the second sequence as key and value, ii) computing multi-head attention of the second sequence as query, and first sequence as key and value, iii) concatenating the attention outputs along feature axis to obtain final sequence output. The same is depicted in Fig. \ref{fig:symmetric_attention}.

We apply \emph{co-attention} (i.e. symmetric attention between sequences of \emph{same} modalities) between $Q_v$ \& $T_v$, $Q_t$ \& $T_t$ and $Q_s$ \& $T_s$ to generate output sequences $VV$, $TT$ and $SS$ respectively. $VV$, $TT$ and $SS$ contain information about the correlation between the query and target features of the same modality.

Building on our starting intuition regarding interconnection between different modalities, we first compute \emph{cross-attention} (i.e. symmetric attention between sequences of \emph{different} modalities) between $Q_s$ -- $T_v$ and $T_s$ -- $Q_v$ to generate \emph{spatio-visual} embeddings $S_q V_t$ and $S_t V_q$. As mentioned earlier, $T_v$ contains information about $T_t$ and likewise, same is the case with $Q_v$ and $Q_t$. Therefore, to leverage these relations, we compute cross-attention of $S_q V_t$ and $S_t V_q$ with $T_t$ and $Q_t$ respectively. Finally, we get \textit{spatio-visio-textual} encodings $S_q V_t T_t$ and $S_t V_q T_q$.

\noindent\textbf{Detection of Similar Snippets.} Finally, we have 5 token sequences (each with max length and feature dimension as 1024) -- 3 Co-Attention sequences: $VV$, $SS$, $TT$ and 2 Cross-Attention sequences: $S_q V_t T_t$ and $S_t V_q T_q$. These sequences are simply concatenated along the last dimension to form a feature volume $F_{sim} \in \mathcal{R}^{BS\times1024\times5120}$, where $BS$ represents the size of the batch and $1024$ is the maximum sequence length (hyperparameter). We posit that this feature volume contains all the necessary information to find the relevant snippets within the target. We apply a linear projection on $F_{sim}$ and convert it to a vector of shape $BS\times 1024\times4096$, that is reshaped into a feature volume $F_{feat} \in \mathcal{R}^{BS\times 1024\times64\times64}$.

A sequence of conv layers, each with a kernel size of $1$, followed by LeakyReLU activation (slope$=0.1$), processes $F_{feat}$ to output features at $4$ different levels, with shape - $BS\times256\times64\times64$, $BS\times512\times64\times64$, $BS\times1024\times64\times64$, $BS\times2048\times64\times64$. The hierarchical features are subsequently processed through a standard FPN architecture, followed by the FasterRCNN  RPN and RoI heads \cite{ren2015faster} to obtain the final bounding boxes. 
Please refer to the supplementary for further details about the FPN and RPN modules, hidden dimension of other modules, size of intermediate vectors obtained through attention,  etc.




\begin{table*}[t]
\centering
\resizebox{470pt}{!}{%
\begin{tabular}{|l|lllll|lllll|}
\hline
\multicolumn{1}{|c|}{\multirow{2}{*}{\textbf{Model}}} & \multicolumn{5}{c|}{\textit{Flamingo Forms}} & \multicolumn{5}{c|}{PubLayNet Documents} \\ \cline{2-11} 
\multicolumn{1}{|c|}{} & \multicolumn{1}{c|}{AP50} & \multicolumn{1}{c|}{AP75} & \multicolumn{1}{c|}{AR50} & \multicolumn{1}{c|}{AR75} & \multicolumn{1}{c|}{mAP} & \multicolumn{1}{c|}{AP50} & \multicolumn{1}{c|}{AP75} & \multicolumn{1}{c|}{AR50} & \multicolumn{1}{c|}{AR75} & \multicolumn{1}{c|}{mAP} \\ \hline
SSD & \multicolumn{1}{l|}{\textbf{94.16}} & \multicolumn{1}{l|}{\textbf{94.16}} & \multicolumn{1}{l|}{0.0000} & \multicolumn{1}{l|}{0.00} & 0.00 & \multicolumn{1}{l|}{99.07} & \multicolumn{1}{l|}{99.06} & \multicolumn{1}{l|}{0.01} & \multicolumn{1}{l|}{0.00} & 0.00 \\
NCC & \multicolumn{1}{l|}{29.41} & \multicolumn{1}{l|}{24.82} & \multicolumn{1}{l|}{5.16} & \multicolumn{1}{l|}{0.00} & 2.77 & \multicolumn{1}{l|}{46.09} & \multicolumn{1}{l|}{29.94} & \multicolumn{1}{l|}{18.60} & \multicolumn{1}{l|}{0.04} & 7.36 \\
BHRL (CVPR'22) & \multicolumn{1}{l|}{58.09} & \multicolumn{1}{l|}{51.00} & \multicolumn{1}{l|}{38.67} & \multicolumn{1}{l|}{30.28} & 35.45 & \multicolumn{1}{l|}{36.74} & \multicolumn{1}{l|}{26.18} & \multicolumn{1}{l|}{54.55} & \multicolumn{1}{l|}{28.69} & 22.47 \\
LayoutLMv3 (MM'22) & \multicolumn{1}{l|}{51.45} & \multicolumn{1}{l|}{43.21} & \multicolumn{1}{l|}{\textbf{58.88}} & \multicolumn{1}{l|}{38.80} & 45.51 & \multicolumn{1}{l|}{35.95} & \multicolumn{1}{l|}{16.50} & \multicolumn{1}{l|}{65.38} & \multicolumn{1}{l|}{18.31} & 21.46 \\ \hline
MONOMER (Ours) & \multicolumn{1}{l|}{78.16} & \multicolumn{1}{l|}{73.93} & \multicolumn{1}{l|}{56.65} & \multicolumn{1}{l|}{\textbf{51.11}} & \textbf{66.95} & \multicolumn{1}{l|}{\textbf{64.30}} & \multicolumn{1}{l|}{\textbf{39.83}} & \multicolumn{1}{l|}{\textbf{64.18}} & \multicolumn{1}{l|}{\textbf{32.95}} & \textbf{36.61} \\ \hline
\end{tabular}}
\caption{Comparing MONOMER's performance against other approaches at the task of one-shot doc snippet detection. (Note: high precision in SSD is a result of no boxes being detected in most of the cases reflected in the mAP and recall.)}
\label{tab:comparison_results}
\end{table*}

\begin{table*}[h]
    \centering
    \resizebox{470pt}{!}{%
   \begin{tabular}{|l|ccccc|ccccc|}
    \hline
    \multicolumn{1}{|c|}{\multirow{2}{*}{\textbf{MONOMER Variant}}} & \multicolumn{5}{c|}{\textit{Flamingo Forms}} & \multicolumn{5}{c|}{PubLayNet Documents} \\ \cline{2-11} 
    \multicolumn{1}{|c|}{} & \multicolumn{1}{c|}{AP50} & \multicolumn{1}{c|}{AP75} & \multicolumn{1}{c|}{AR50} & \multicolumn{1}{c|}{AR75} & mAP & \multicolumn{1}{c|}{AP50} & \multicolumn{1}{c|}{AP75} & \multicolumn{1}{c|}{AR50} & \multicolumn{1}{c|}{AR75} & mAP \\ \hline
    Image & \multicolumn{1}{c|}{67.33} & \multicolumn{1}{c|}{62.40} & \multicolumn{1}{c|}{\textbf{59.49}} & \multicolumn{1}{c|}{49.95} & 63.73 & \multicolumn{1}{c|}{53.43} & \multicolumn{1}{c|}{30.13} & \multicolumn{1}{c|}{60.29} & \multicolumn{1}{c|}{23.98} & 23.75 \\
    Image + Text & \multicolumn{1}{c|}{72.46} & \multicolumn{1}{c|}{67.60} & \multicolumn{1}{c|}{57.97} & \multicolumn{1}{c|}{50.25} & 64.31 & \multicolumn{1}{c|}{62.08} & \multicolumn{1}{c|}{36.53} & \multicolumn{1}{c|}{58.03} & \multicolumn{1}{c|}{27.27} & 33.00 \\
    Image + Bounds & \multicolumn{1}{c|}{70.37} & \multicolumn{1}{c|}{65.50} & \multicolumn{1}{c|}{57.30} & \multicolumn{1}{c|}{49.06} & 63.30 & \multicolumn{1}{c|}{57.67} & \multicolumn{1}{c|}{34.10} & \multicolumn{1}{c|}{\textbf{69.21}} & \multicolumn{1}{c|}{32.33} & 32.91 \\ \hline
    Image + Text + Bounds & \multicolumn{1}{c|}{\textbf{78.16}} & \multicolumn{1}{c|}{\textbf{73.93}} & \multicolumn{1}{c|}{56.65} & \multicolumn{1}{c|}{\textbf{51.11}} & \textbf{66.95} & \multicolumn{1}{c|}{\textbf{64.30}} & \multicolumn{1}{c|}{\textbf{39.83}} & \multicolumn{1}{c|}{64.18} & \multicolumn{1}{c|}{\textbf{32.95}} & \textbf{36.61} \\ \hline
    \end{tabular}}
    \caption{Analysing performance of MONOMER variants that use different combinations of modalities.}
    \label{tab:ablation_results}
\end{table*}

\section{Experiments and Analysis}



\subsection{Implementation Details} We train MONOMER using standard Object Detection losses i.e proposal matching + bounding box (used in Faster-RCNN) \cite{ren2015faster} on a batch size of 48 (6 per GPU, total of 8 GPUs). Optimization is performed using SGD \cite{ruder2016overview} with momentum 0.9 and weight decay 1e-2. The initial learning rate is set to 5e-2 and updated with a cosine annealing scheduler. The output of detection head is processed with a confidence threshold of 0.4 on prediction and NMS \cite{neubeck2006efficient} threshold of 0.45 on IoU. For all the experiments, we uniformly use 8 Nvidia A-100 GPUs.

\subsection{Baselines}
We begin with applying standard template matching approaches: Normalized Cross-Correlation (NCC) and Sum of Squared Differences (SSD) to detect similar snippets. Further, owing to resemblance of the proposed task's with one-shot object detection (OSOD) setting, we train BHRL\footnote{\url{https://github.com/hero-y/BHRL}}, the current state-of-the-art in OSOD. Additionally, we implement an approach using top-performing document analysis model LayoutLMv3\footnote{\url{https://huggingface.co/docs/transformers/model\_doc/layoutlmv3}}, where the query-target are embedded separately to generate multi-modal features that are processed through symmetric attention and detection head.

\subsection{Evaluation Metrics} We adopt the metrics from one-shot object detection for evaluating performances of various approaches on the one-shot doc snippet detection. Specifically, we measure Average Precision(AP) and Average Recall(AR) at IoU thresholds of 0.50 and 0.75 which are denoted by AP$_{50}$, AP$_{75}$, AR$_{50}$ and AR$_{75}$ respectively. In addition, we calculate mean Average Precision (mAP) \cite{lin2014microsoft} of the predictions by averaging APs at IoU thresholds starting from 0.50 and increasing in the steps of 0.05 till 0.95.

\subsection{Results} 
\label{sec:comp_baselines}
Table \ref{tab:comparison_results} shows the results of different approaches at doc snippet detection task. We see that the template matching algorithms perform very poorly on this task, the reason being their inability to  adapt to the transformations in the similar snippets such as aspect ratios, font sizes, styles, and the like. BHRL shows significant improvement over template matching, but its performance plateaus early because of its lack of understanding of the text and layout information in the documents. LayoutLMv3, with its rich document representations, demonstrates improvement over aforementioned techniques. Nevertheless, using multi-modal embeddings directly as in the LayoutLMv3's extension, doesn't provide an explicit control over individual modalities of the query and the target. The effective outcome is, MONOMER, with more flexibility to process information streams, gives better mAP in both the data settings.

\begin{figure*}
    \centering
    \includegraphics[width=\linewidth]{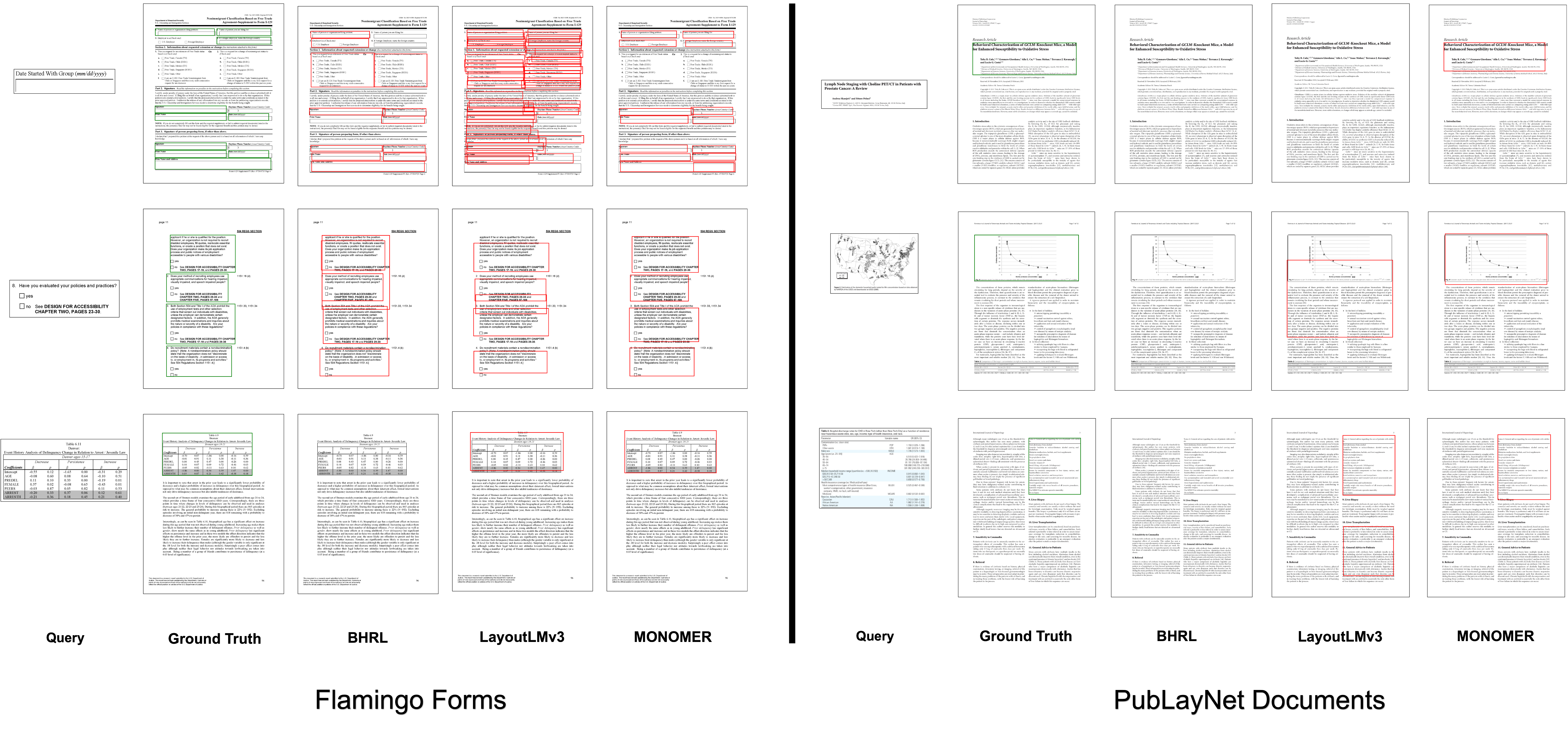}
    \caption{Qualitative comparison between BHRL, LayoutLMv3 and the proposed MONOMER. Please use high levels of \textbf{zooming in} to comprehend the visualization.}
    \label{fig:vis_forms_plus_pub}
\end{figure*}

\noindent \textbf{Qualitative Visualizations.} 
In this section, we discuss the key differences between the qualitative outputs produced by MONOMER against other strong baselines. A summary of the results is illustrated in ~Fig.  \ref{fig:vis_forms_plus_pub}. The query snippets contain certain layout patterns whose corresponding matches in the target document are shown by the green bounds in ground truth columns. As we can observe, the query is not \textit{exactly} the same as regions marked in the ground truth, thus making the detection task non-trivial. We note that MONOMER is able to detect several complex patterns in the form, clearly performing better than the detections made by the baseline approaches. For instance, the top-left (row 1, Flamingo-Forms) example in ~Fig. \ref{fig:vis_forms_plus_pub} demonstrates that while BHRL is able to detect most of the true positives, it also detects two regions as false positives. We attribute this behavior of BHRL on its over-reliance on a limited number of classes (all choice-like patterns are detected instead of the similar layout). Further, LayoutLMv3 also predicts a number of extraneous bounding boxes that do not match the ground truth. Similarly, at the bottom-left in ~Fig. \ref{fig:vis_forms_plus_pub} (row 3, Flamingo Forms) the superior precision of MONOMER over both LayoutLMv3 and BHRL can be observed. Furthermore, MONOMER yields better quality detections even in the PubLayNet dataset as shown in Fig. \ref{fig:vis_forms_plus_pub} (right). We note both BHRL and LayoutLMv3 often fail to predict bounding boxes for examples in the PubLayNet dataset, whereas MONOMER consistently predicts the expected bounding boxes. The efficacy of our method over LayoutLMv3 and BHRL can be observed in Fig. \ref{fig:vis_forms_plus_pub} (row 3, PubLayNet) wherein LayoutLMv3 produces a false positive and BHRL does not yield any prediction. Our qualitative observations are consistent with the quantitative results shown in Table \ref{tab:comparison_results}. Please refer to supplementary for more qualitative analysis.

\subsection{Ablation And Analysis}
\label{sec:ablation}
\noindent\textbf{Performance on varying Modalities.} We quantify the roles played by individual document modalities in MONOMER's performance through an ablation study where we switch the modality information on/off in different combinations. First, we consider image-only variant of MONOMER. To this model, we add text and bounding box modalities separately to get two more MONOMER variants. Table \ref{tab:ablation_results} compares performances of these variants against MONOMER trained on image, text and bounds together. Model processing all modalities surpasses other variants significantly, underlining the usefulness of incorporating document-specific nuances in the architecture.

\begin{table}[t]
    \centering
    \begin{tabular}{|l|ll|ll|}
    \hline
    \multicolumn{1}{|c|}{\multirow{2}{*}{\textbf{Model}}} & \multicolumn{2}{c|}{\textit{Flamingo}} & \multicolumn{2}{c|}{\textit{PubLayNet}} \\ \cline{2-5} 
    \multicolumn{1}{|c|}{} & \multicolumn{1}{c|}{Seen} & \multicolumn{1}{c|}{Unseen} & \multicolumn{1}{c|}{Seen} & \multicolumn{1}{c|}{Unseen} \\ \hline
    NCC & \multicolumn{1}{l|}{0.00} & 0.00 & \multicolumn{1}{l|}{0.01} & 0.00 \\
    SSD & \multicolumn{1}{l|}{0.81} & 1.95 & \multicolumn{1}{l|}{0.37} & 7.00 \\
    BHRL & \multicolumn{1}{l|}{47.50} & 42.30 & \multicolumn{1}{l|}{16.10} & 16.00 \\
    LayoutLMv3 & \multicolumn{1}{l|}{53.98} & 42.03 & \multicolumn{1}{l|}{24.48} & 18.44 \\ \hline
    MONOMER & \multicolumn{1}{l|}{\textbf{71.33}} & \textbf{57.82} & \multicolumn{1}{l|}{\textbf{31.86}} & \textbf{31.27} \\ \hline
    \end{tabular}
    \caption{Study of generalization capabilities of various approaches in one-shot setting (numbers in mAP).}
    \label{tab:flm_plus_pub_seen_unseen}
\end{table}

\noindent\textbf{Performance on Unseen Layout Strings.} Now, we evaluate various approaches for their ability to detect snippet patterns that were not encountered by the approach during its training. This would test one-shot detection capabilities of the approaches. We distinguish  seen-unseen classes by checking whether a layout string pattern in testset appeared in the trainset or not. The testset for Flamingo contains 1558 seen layout patterns and 353 unseen layout patterns; similarly, PubLayNet testset comprises of 17 seen and 6 unseen layout patterns. When inference is performed separately on the seen-unseen split, we obtain results as shown in table \ref{tab:flm_plus_pub_seen_unseen}. The numbers depict MONOMER's superiority over other approaches in  correctly identifying unseen layout strings, and thus, underscore its efficiency in inferring layout strings of even the unseen snippet patterns.

\section{Conclusion and Future Work}

In this work, we explore a multi-modal one-shot detection setting for enhancing search within documents. Discussing the similarity in the context of documents, we propose a similarity criterion that allows generation of large amount of data required for testing out different approaches. Then, we propose a cross-attention based solution that is built upon insights into how various document modalities for queried snippet and target documents are inter-related. Our approach shows better performance compared to other approaches and its own single modality variants for the task of one-shot document snippet detection. 

In future, we wish to extend this work to other multi-modal content such as infographics, advertisement flyers, handwritten text, etc. Also, we believe that the search paradigm described in this work could be further augmented with textual intents (in addition to text contained within the document) for the search. This would allow users to perform a search by providing an intent and an example document snippet. Likewise, this work opens up discussion on many possibilities for search in documents.

\section{Acknowledgement}
We would like to sincerely thank Arneh Jain, Mayur Hemani and Hiresh Gupta for helpful discussions and feedback that helped inform this work.
\section{Appendix}
\appendix
\section{Additional Architectural Details}
\label{appendix:add_architecture}
\subsection{Feature Extractors.}
In this section, we provide additional details about the backbones in the proposed MONOMER. We discuss the hyperparameter choices in the \textit{Image Encoder}, followed by the \textit{Text Encoder} and finally the \textit{Layout Encoder}. Please note that the hyperparameters are kept consistent across experiments on all the datasets.
\subsubsection{Image Features}
The image encoder is a DiT-backbone with encoder-only architecture having $4$ layers, each containing $4$ attention heads with model dimension of $512$. The encoder takes $3$ channel document image resized (using bi-cubic interpolation) to $224\times224$ resolution which is further cut into $16\times16$ sized patches and outputs a token sequence of length $197$. The $197$ tokens are formed as follows -- $\frac{224\times224}{16\times16} + 1$, where the additional token corresponds to the $CLS$ token as in the original BEiT \cite{bao2021beit}. We choose a pretrained DiT base model for our experiments that has a hidden dimension of $768$. Since both query image $Q_{i}^{inp}$ and target image $T_{i}^{inp}$ are preprocessed to the same dimension, we obtain two feature vectors $Q_{v}$, $T_{v}$ each of size $BS\times197\times1024$, where 1024 is the maximum sequence length and $BS$ denotes the batch size. Note that the maximum sequence length is a hyperparameter choice that is chosen based on the maximum number of text-blocks in the target document. The encodings are then padded to final vectors $Q_{v}, T_{v}$ of size $BS\times1024\times1024$ each. The rationale behind doing so is to conveniently be able to perform the subsequent cross-attention with different modalities. We can summarize the sequence of operations as follows - 
\begin{gather}
    Q_{v} = pad(D(Q_{i}^{inp})) \in \mathcal{R}^{1024\times1024} \\
    T_{v} = pad(D(T_{i}^{inp})) \in \mathcal{R}^{1024\times1024}
\end{gather}

where $D$ is the DiT image encoder and $pad$ is the padding operation.

\subsubsection{Text Features}
We use a pretrained BeRT-based sentence transformer \cite{song2020mpnet} that generates 768 dimensional embedding for a given block of text. The continuous blocks of text in the document are fed into this encoder to generate token sequence $T_{t}^{inp}$, $Q_{t}^{inp}$ of dimension $BS\times text_{t} \times 768$ and $BS\times text_q \times 768$ respectively, where $text_{t}$ is the number of text-blocks in the target document and $text_q$ is the number of text-blocks in the query patch. Additionally, we pad both $T_{t}^{inp}$, $Q_{t}^{inp}$ to a constant size of $BS\times1024\times768$. Unlike all the other MONOMER parameters, the text encoder weights are kept \emph{frozen}. Mathematically, text encoding is represented as follows --
\begin{gather}
    Q_{t} = B(pad(Q_{t}^{inp})) \in \mathcal{R}^{1024\times768} \\
    T_{t} = B(pad(T_{t}^{inp})) \in \mathcal{R}^{1024\times768}
\end{gather}
where $Q_{t}, T_{t}$ are the final query and target text features and $B$ is the BeRT text encoder.

\begin{table*}[h]
\centering
\begin{tabular}{|l|l|l|l|l|c|}
\hline
\textbf{Metrics} & \textbf{Split-1} & \textbf{Split-2} & \textbf{Split-3} & \textbf{Split-4} & \textbf{Average over Splits} \\ \midrule
\textit{Recall} & $87.25$ & $71.93$ & $80.74$ & $84.38$ & $81.07$ \\ \hline
\textit{Precision} & $83.06$ & $87.83$ & $91.41$ & $89.58$ & $87.96$ \\ \hline
\textit{F1} & $84.28$ & $78.56$ & $85.55$ & $86.48$ & $83.71$ \\ \hline
\textit{\begin{tabular}[c]{@{}l@{}}\% Complex\\ Pattern\end{tabular}} & $43.33$ & $39.16$ & $80.0 $& $30.0 $& $48.12$ \\ \hline
\textit{\begin{tabular}[c]{@{}l@{}}\% non exact but similar matches over correct \\ matches highlighted.\end{tabular}} & $79.61$ & $93.07$ & $91.09$ & $86.17$ & $87.48$ \\ \hline
\end{tabular}
\caption{Human Evaluation of the proposed dataset over 4 different splits. The last row indicates the non-trivial nature of dataset generation through the percentage of examples that are not exactly the same as the query.} \label{tab:eval}
\end{table*}

\subsubsection{Bounding Box Features}
We leverage a ViT-like \cite{vaswani2017attention} architecture to encode the bounding box (spatial) information in the target document and query patch. We implement an encoder-only transformer architecture with $4$ layers, $4$ heads and hidden dimension of $1024$. It takes bounds of the target $T_{s}^{inp}$ and query $Q_{s}^{inp}$ of size $BS\times box_{t}\times 4$, $BS\times box_{q}\times 4$, where $box_{t}, box_{q}$ are the number of bounding boxes in target and query respectively. Similar to the text-encoder, $box_{t}$ and $box_{q}$ are padded to the maximum sequence length of $1024$. Weights of this encoder are initialized randomly. We denote the bounding box encoding as follows --
\begin{gather}
    Q_{s} = V(pad(Q_{s}^{inp})) \in \mathcal{R}^{1024\times1024} \\
    T_{s} = V(pad(T_{s}^{inp})) \in \mathcal{R}^{1024\times1024}
\end{gather}
where $V$ is the ViT-like bounding box encoder and $Q_{s}, T_{s}$ are the final feature sets corresponding the query and target respectively.

\subsection{Feature Fusion}
\subsubsection{Symmetric Attention Module.}
The symmetric attention module consists of $2$ multi-head attention (MHA) modules each containing $4$ heads and embedding dimension of $512$. To ensure that the input token feature dimension matches with MHA's specifications, the input sequences are passed through fully-connected layers to project feature dimension onto dimension of $512$. The outputs of the MHA blocks are concatenated (along last dimension) to obtain final token sequence with feature dimension of $BS\times 1024\times1024$. 

\subsubsection{Co-Attention and Cross-Attention Modules.}
Co-Attention Module contains $3$ symmetric attention modules one for each modality, outputting sequences $VV$, $TT$ and $SS$ of length $1024$ and token size $1024$.
\begin{gather}
    VV = SA(Q_{v}, T_{v}) \in \mathcal{R}^{1024\times1024} \\
    TT = SA(Q_{t}, T_{t}) \in \mathcal{R}^{1024\times1024} \\
    SS = SA(Q_{s}, T_{s}) \in \mathcal{R}^{1024\times1024}
\end{gather}

where $SA$ is the Symmetric Attention operation.

Similarly, the Cross-Attention Module consists of 2 symmetric attention modules for generating spatio-visual features and 2 for attending text over those generated features. It generates $S_q V_t T_t$ and $S_t V_q T_q$ the dimensions of which are, once again, length of 1024 and token size of 1024. Finally we concatenate the outputs of Co-Attention and Cross-Attention blocks to create feature volume $F_{sim}$. Obtain $F_{sim}$ as follows --
\begin{gather}
    S_{q}V_{t} = SA(Q_{s}, T_{v}) \in \mathcal{R}^{1024\times1024} \\
    S_{q}V_{t}T_{t} = SA(S_{q}V_{t}, T_{t}) \in \mathcal{R}^{1024\times1024} \\
    S_{t}V_{q} = SA(T_{s}, Q_{v}) \in \mathcal{R}^{1024\times1024} \\
    S_{t}V_{q}T_{q} = SA(S_{t}V_{q}, Q_{t}) \in \mathcal{R}^{1024\times1024} \\
    F_{sim} = concat(VV, TT, SS, S_{q}V_{t}T_{t}, S_{t}V_{q}T_{q})
\end{gather}
where $F_{sim} \in \mathcal{R}^{1024\times1024}$ is the final set of features and are processed as described in the main paper.

\begin{figure*}
\centering
\begin{subfigure}{.48\textwidth}
    \centering
    \includegraphics[width=.98\linewidth]{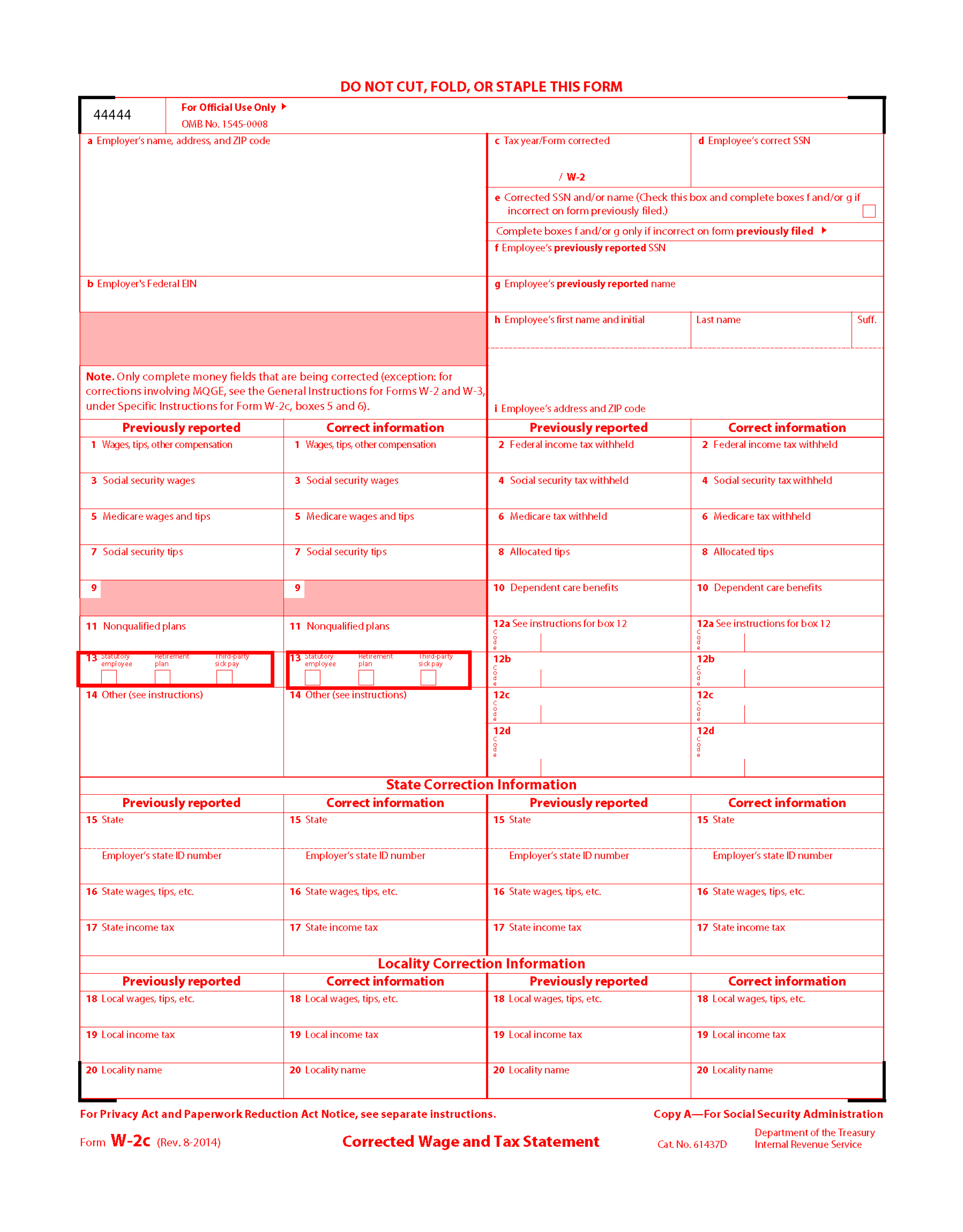}  
    \caption{Pred A}
    \label{p1}
\end{subfigure}
\begin{subfigure}{.48\textwidth}
    \centering
    \includegraphics[width=.98\linewidth]{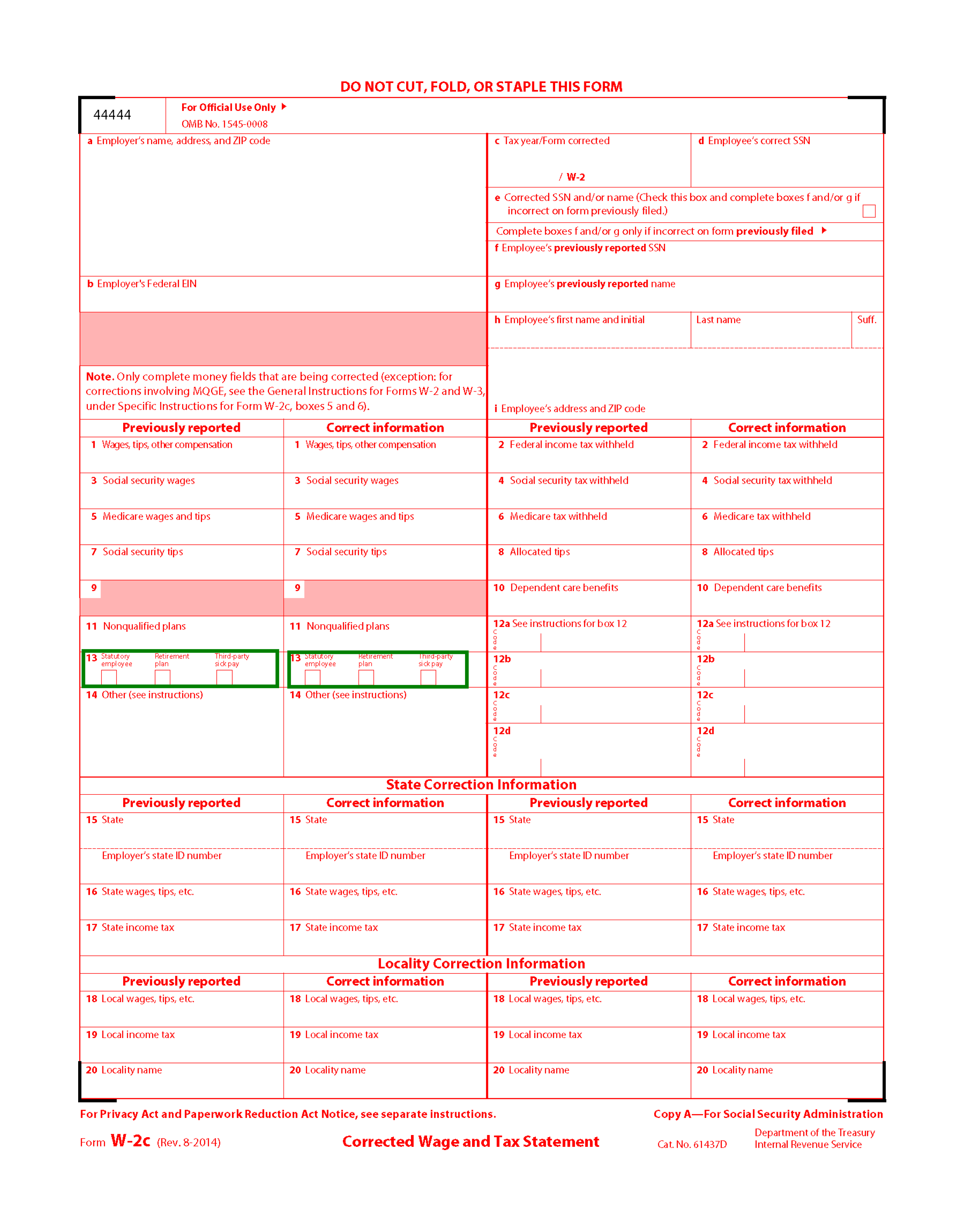}  
    \caption{Target A}
    \label{t1}
\end{subfigure}
\vspace*{1mm}
\begin{subfigure}{.48\textwidth}
    \centering
    \includegraphics[width=.98\linewidth]{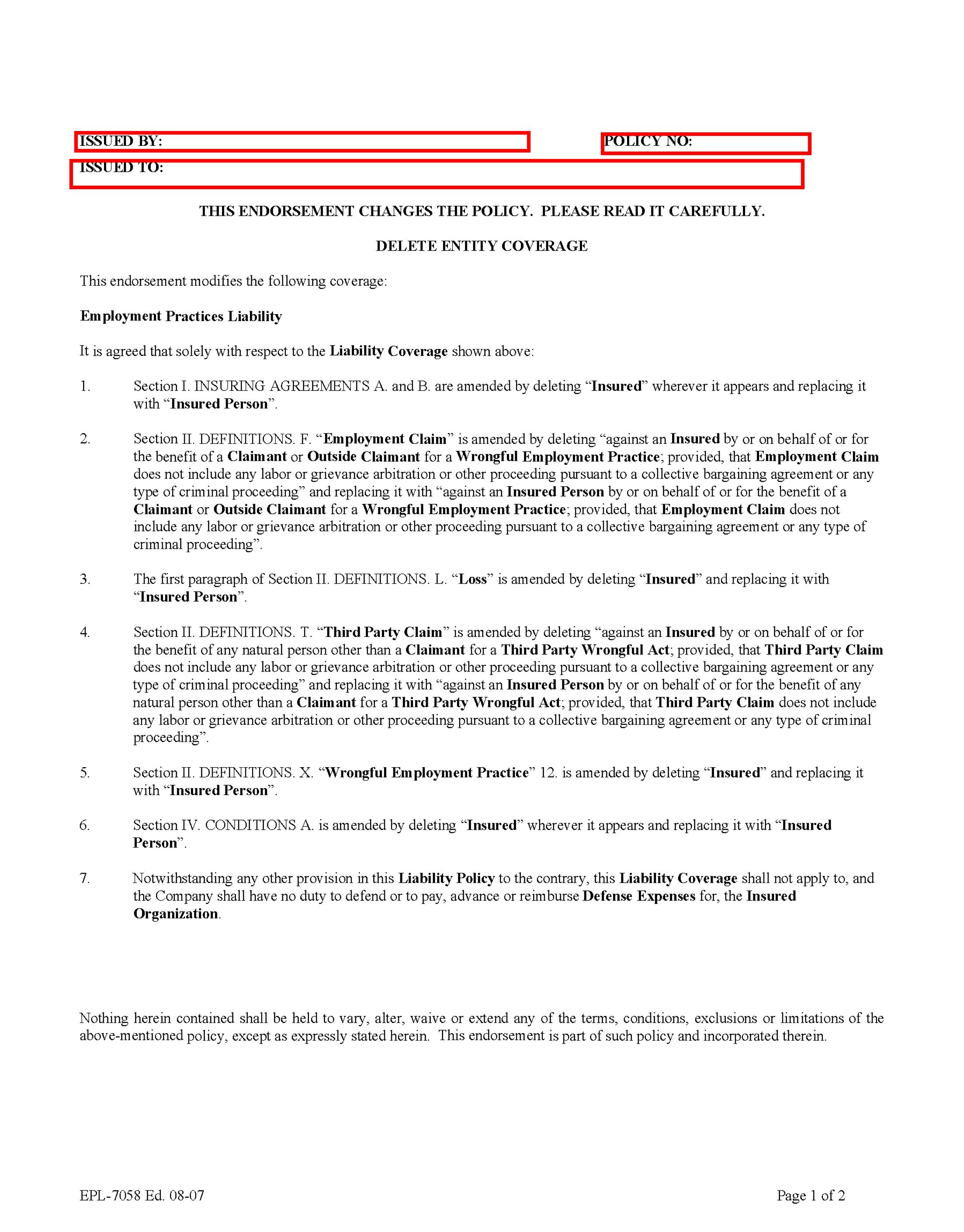}  
    \caption{Pred B}
    \label{p3}
\end{subfigure}
\begin{subfigure}{.48\textwidth}
    \centering
    \includegraphics[width=.98\linewidth]{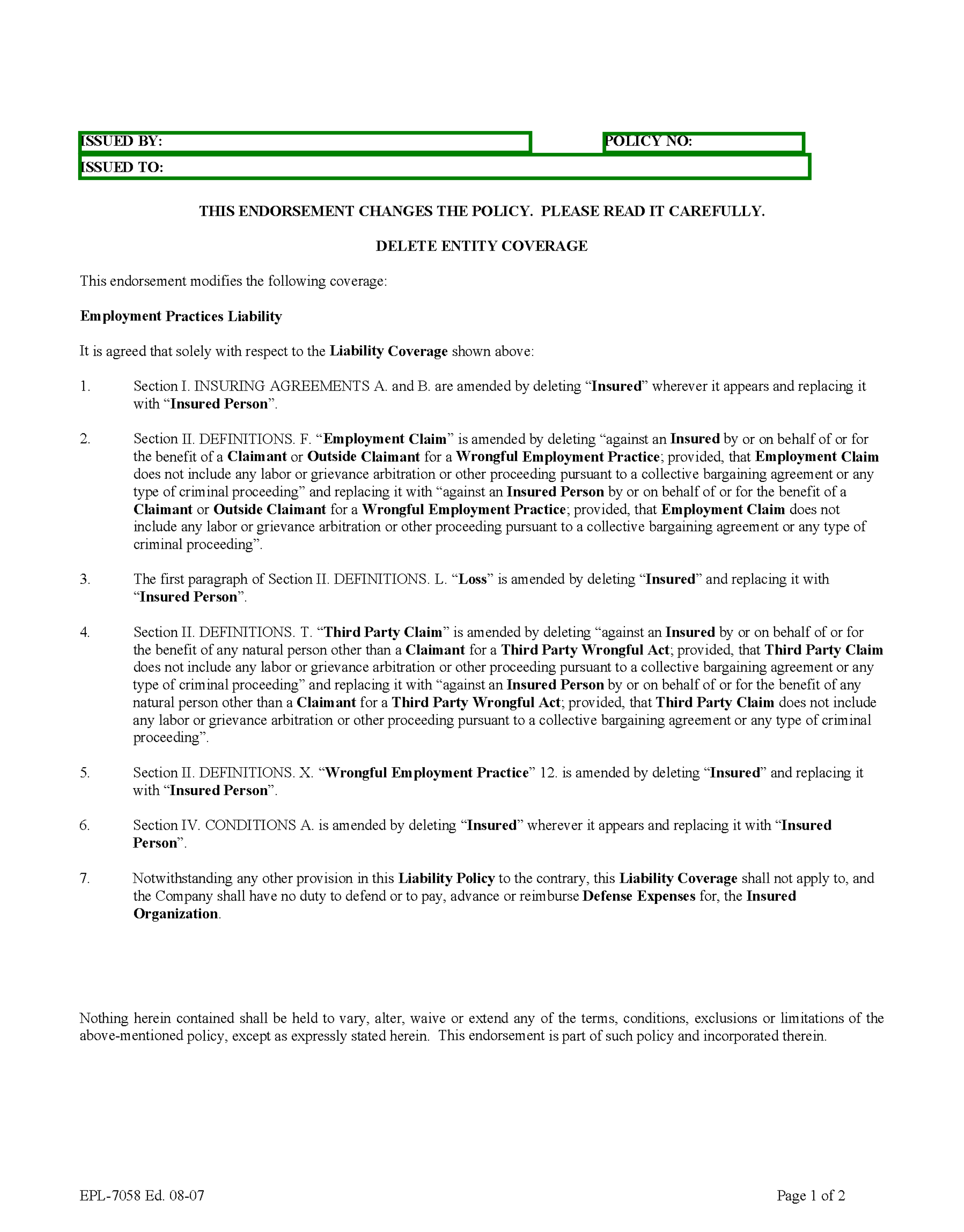}  
    \caption{Target B}
    \label{t3}
\end{subfigure}

\caption{Flamingo Forms Examples (1)}
\label{forms1}
\end{figure*}

\begin{figure*}
\centering

\begin{subfigure}{.48\textwidth}
    \centering
    \includegraphics[width=.98\linewidth]{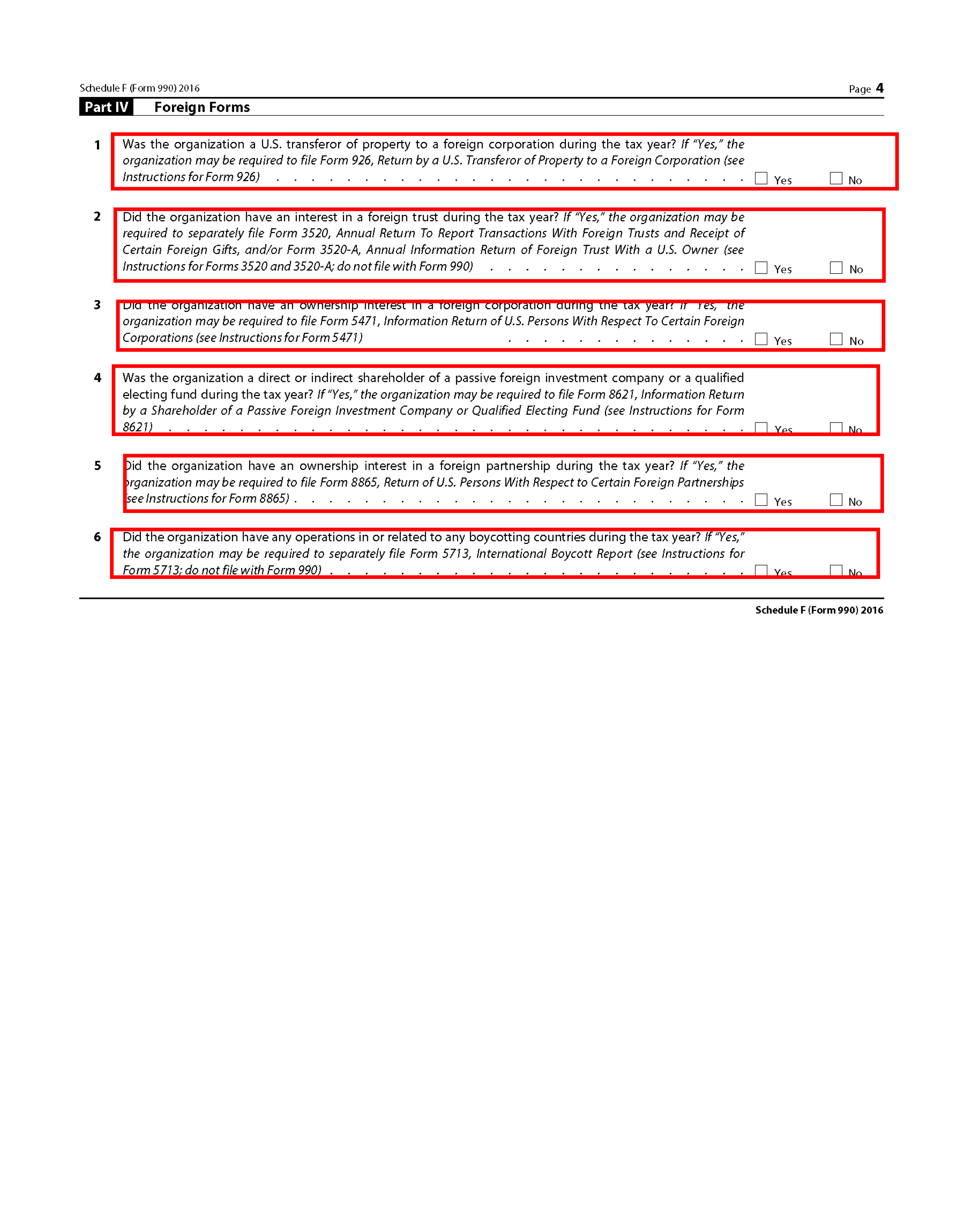}  
    \caption{Pred C}
    \label{p4}
\end{subfigure}
\begin{subfigure}{.48\textwidth}
    \centering
    \includegraphics[width=.98\linewidth]{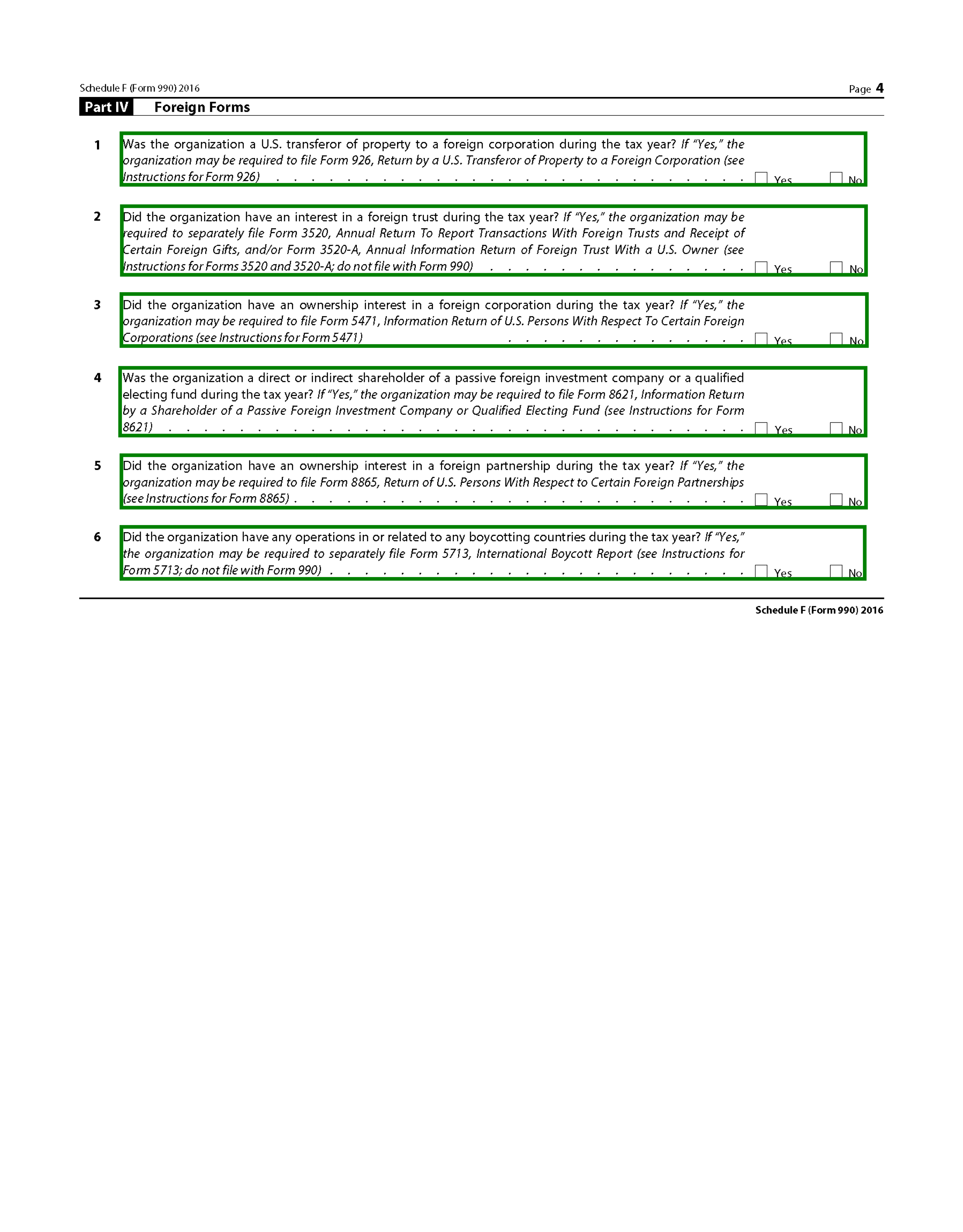}  
    \caption{Target C}
    \label{t4}
\end{subfigure}
\vspace*{2mm}
\begin{subfigure}{.48\textwidth}
    \centering
    \includegraphics[width=.98\linewidth]{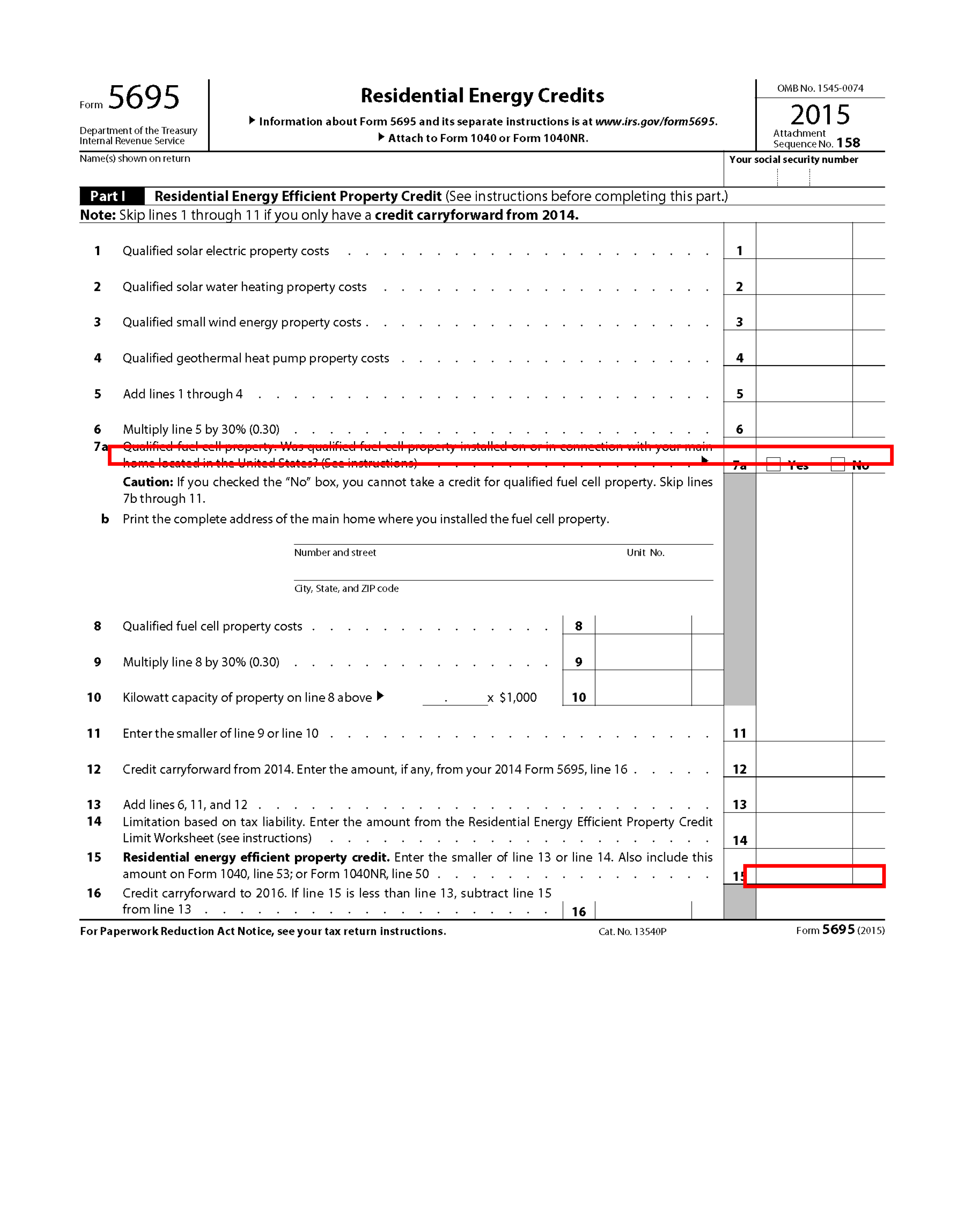}  
    \caption{Pred D}
    \label{p5}
\end{subfigure}
\begin{subfigure}{.48\textwidth}
    \centering
    \includegraphics[width=.98\linewidth]{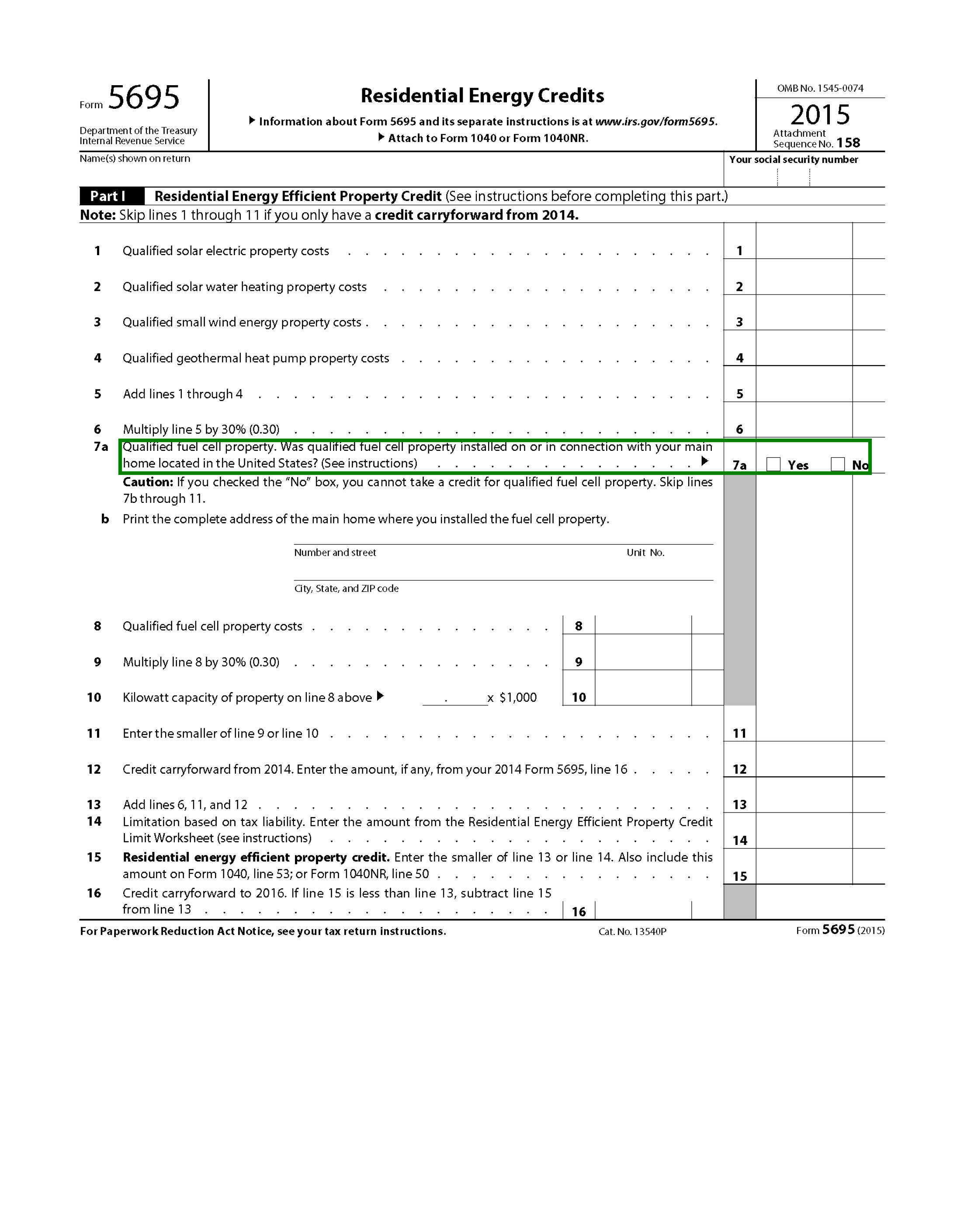}
    \caption{Target D}
    \label{t5}
\end{subfigure}

\caption{Flamingo Forms Examples (2)}
\label{forms2}
\end{figure*}

\begin{figure*}
\centering

\begin{subfigure}{.48\textwidth}
    \centering
    \includegraphics[width=.98\linewidth]{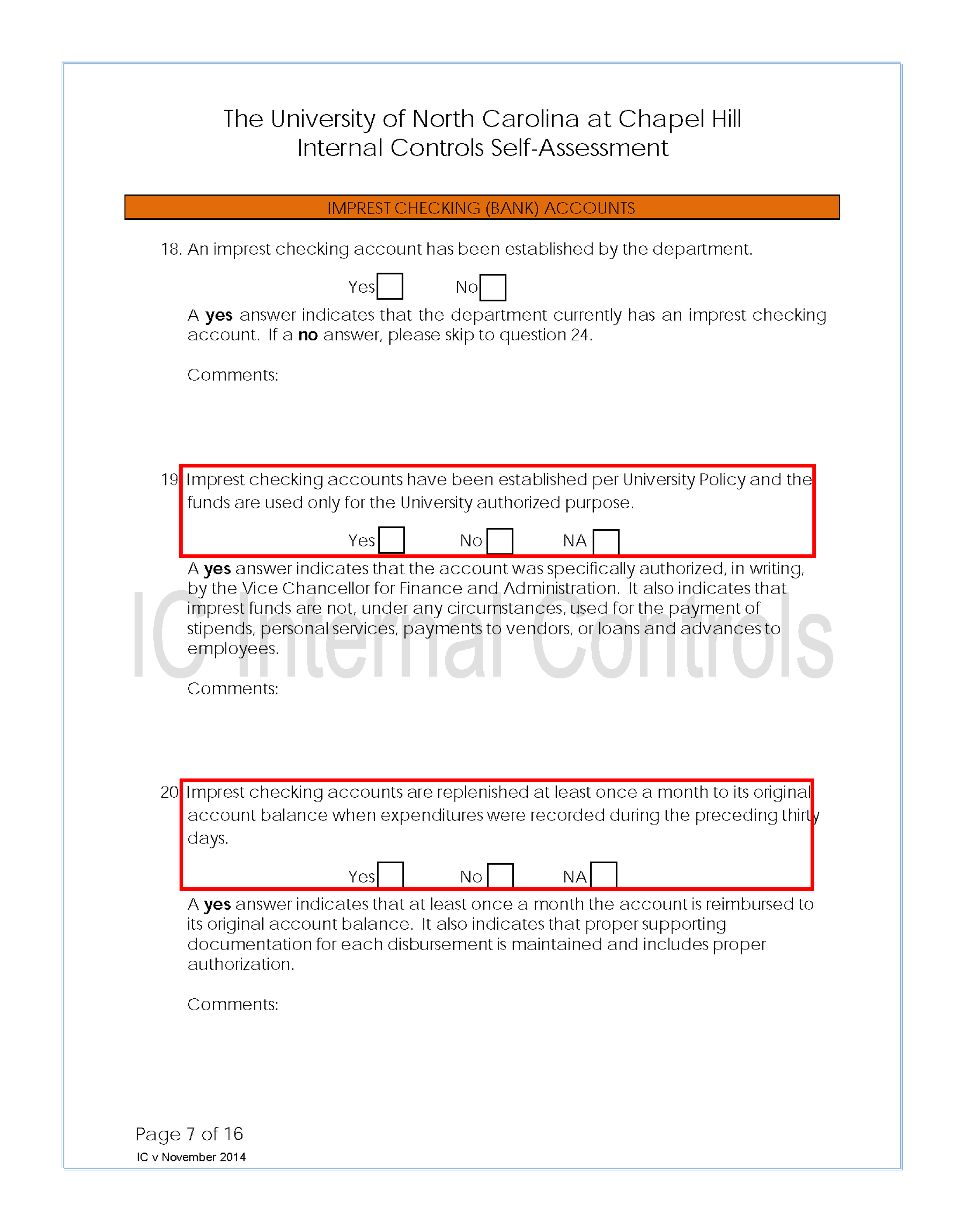}  
    \caption{Pred E}
    \label{p5}
\end{subfigure}
\begin{subfigure}{.48\textwidth}
    \centering
    \includegraphics[width=.98\linewidth]{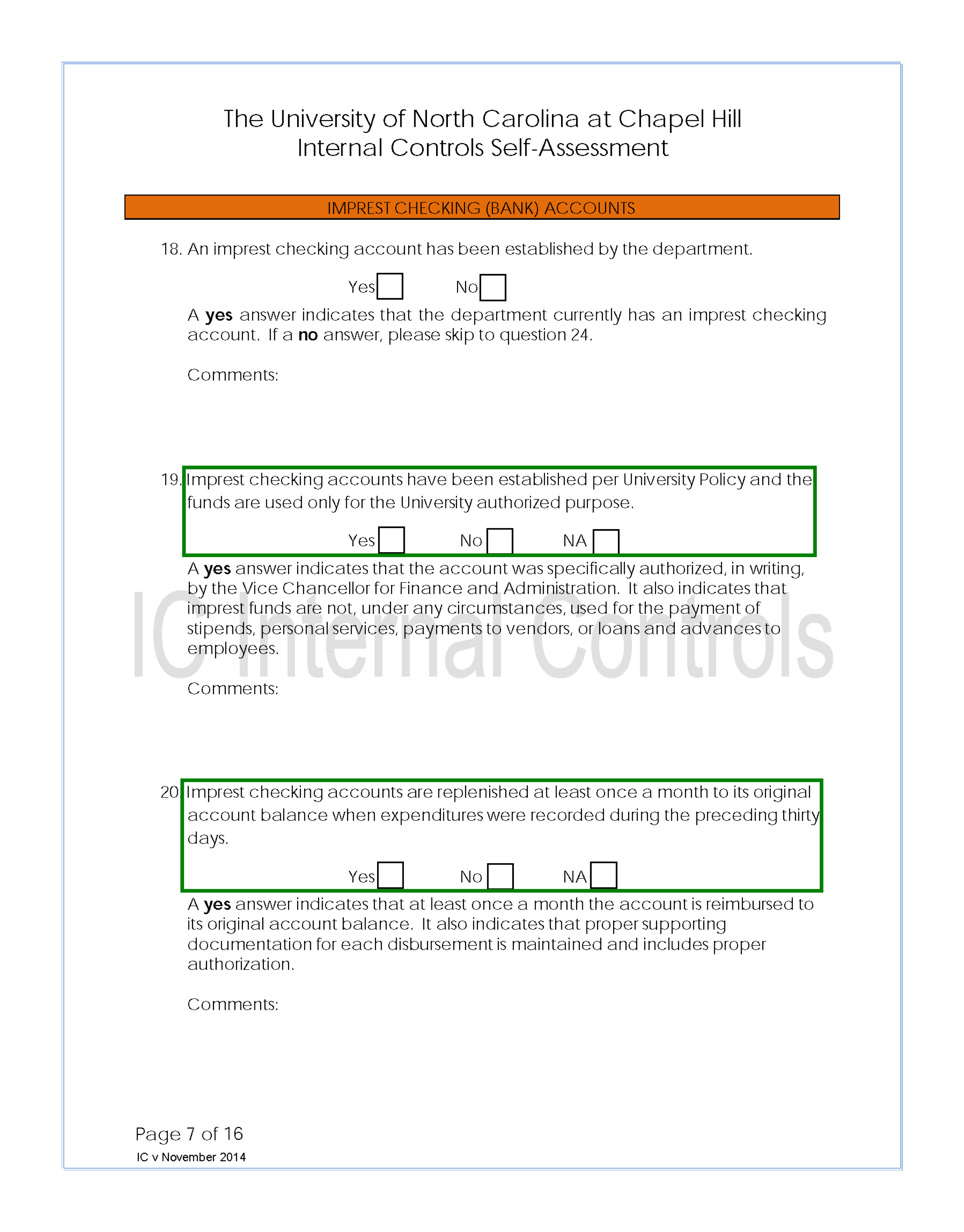}  
    \caption{Target E}
    \label{t5}
\end{subfigure}
\vspace*{2mm}
\begin{subfigure}{.48\textwidth}
    \centering
    \includegraphics[width=.98\linewidth]{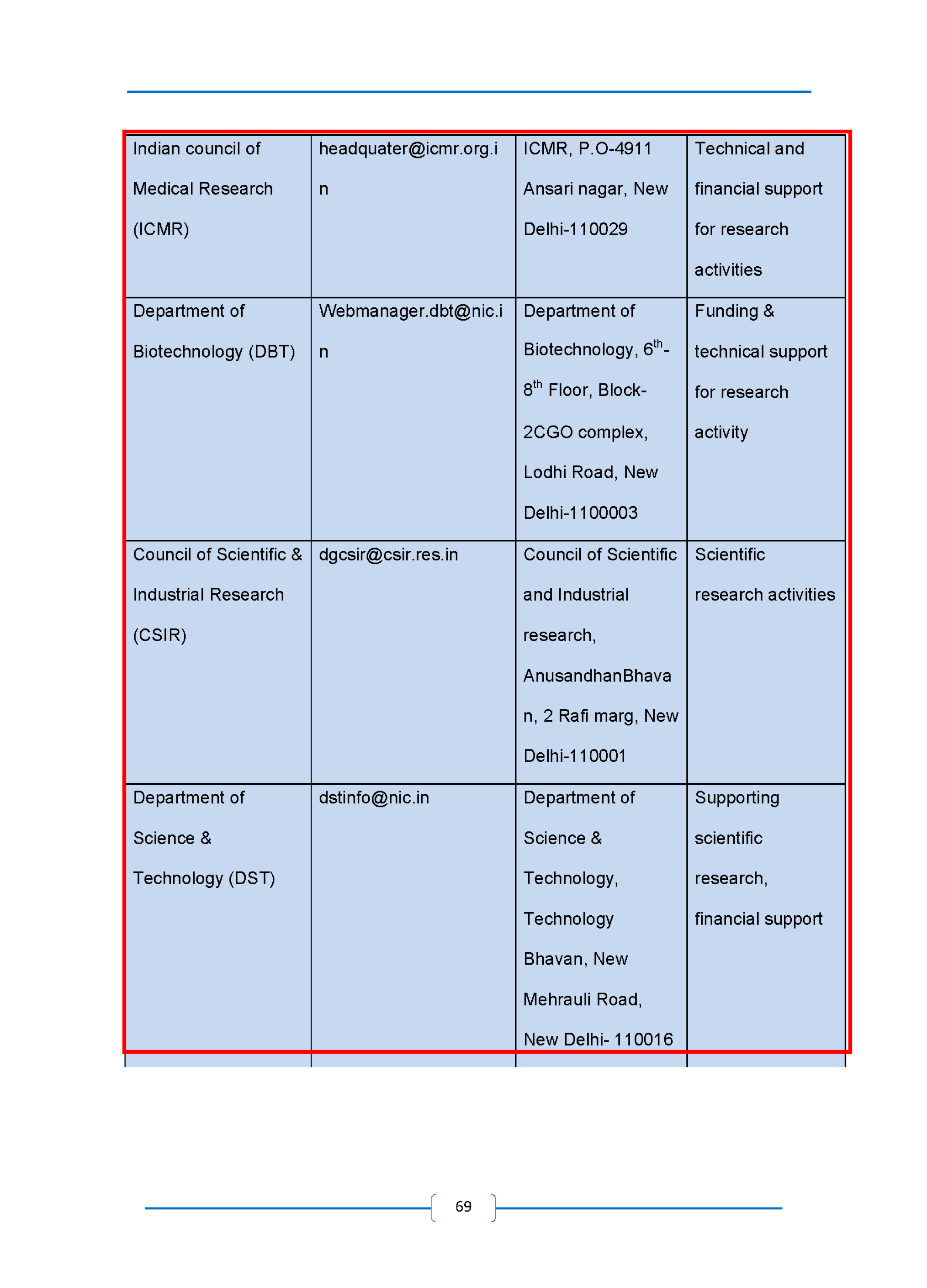}  
    \caption{Pred F}
    \label{p6}
\end{subfigure}
\begin{subfigure}{.48\textwidth}
    \centering
    \includegraphics[width=.98\linewidth]{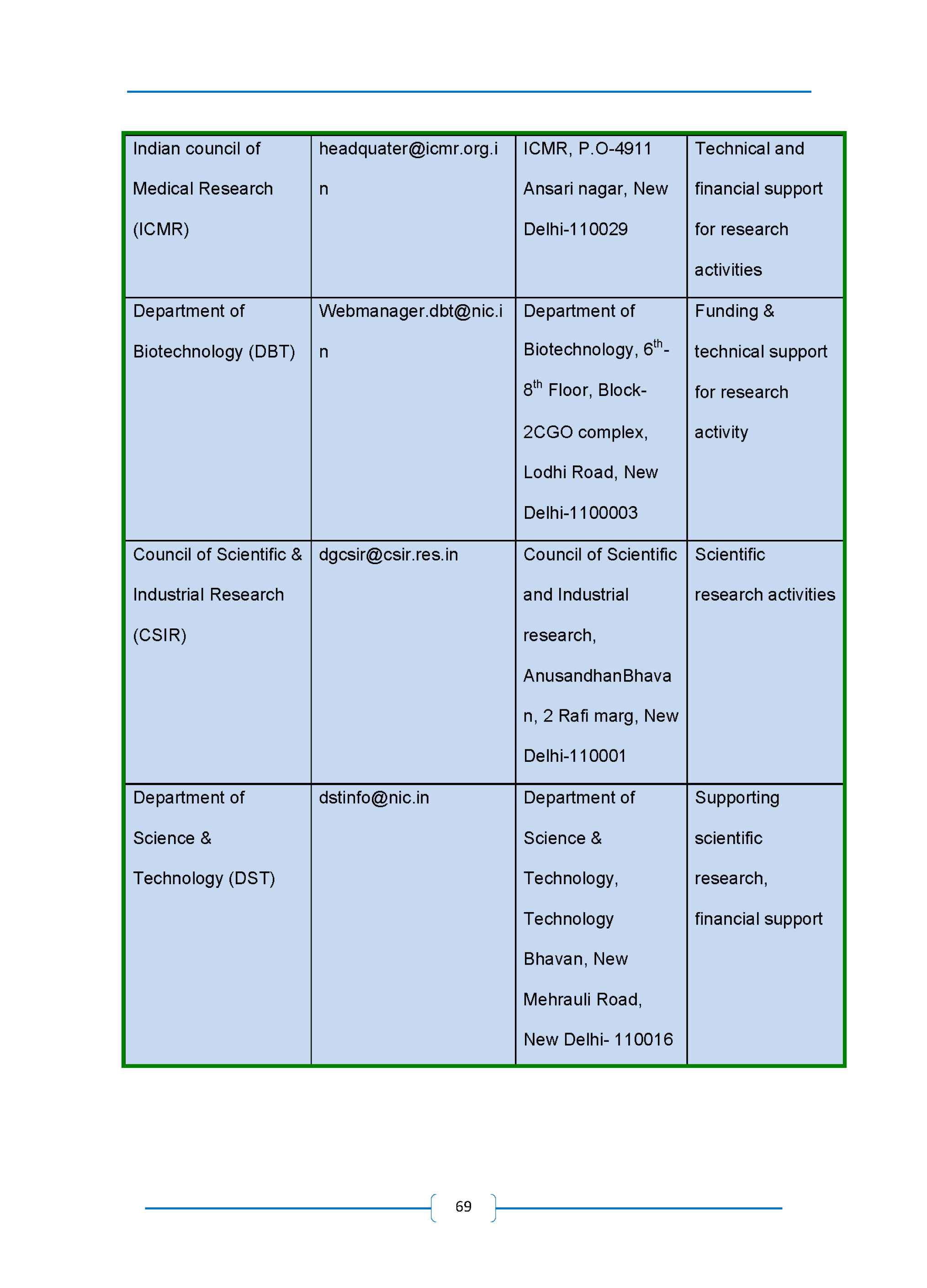}  
    \caption{Target F}
    \label{t6}
\end{subfigure}

\caption{Flamingo Forms Examples (3)}
\label{forms3}
\end{figure*}

\begin{figure*}
\centering

\begin{subfigure}{.47\textwidth}
    \centering
    \includegraphics[width=.98\linewidth]{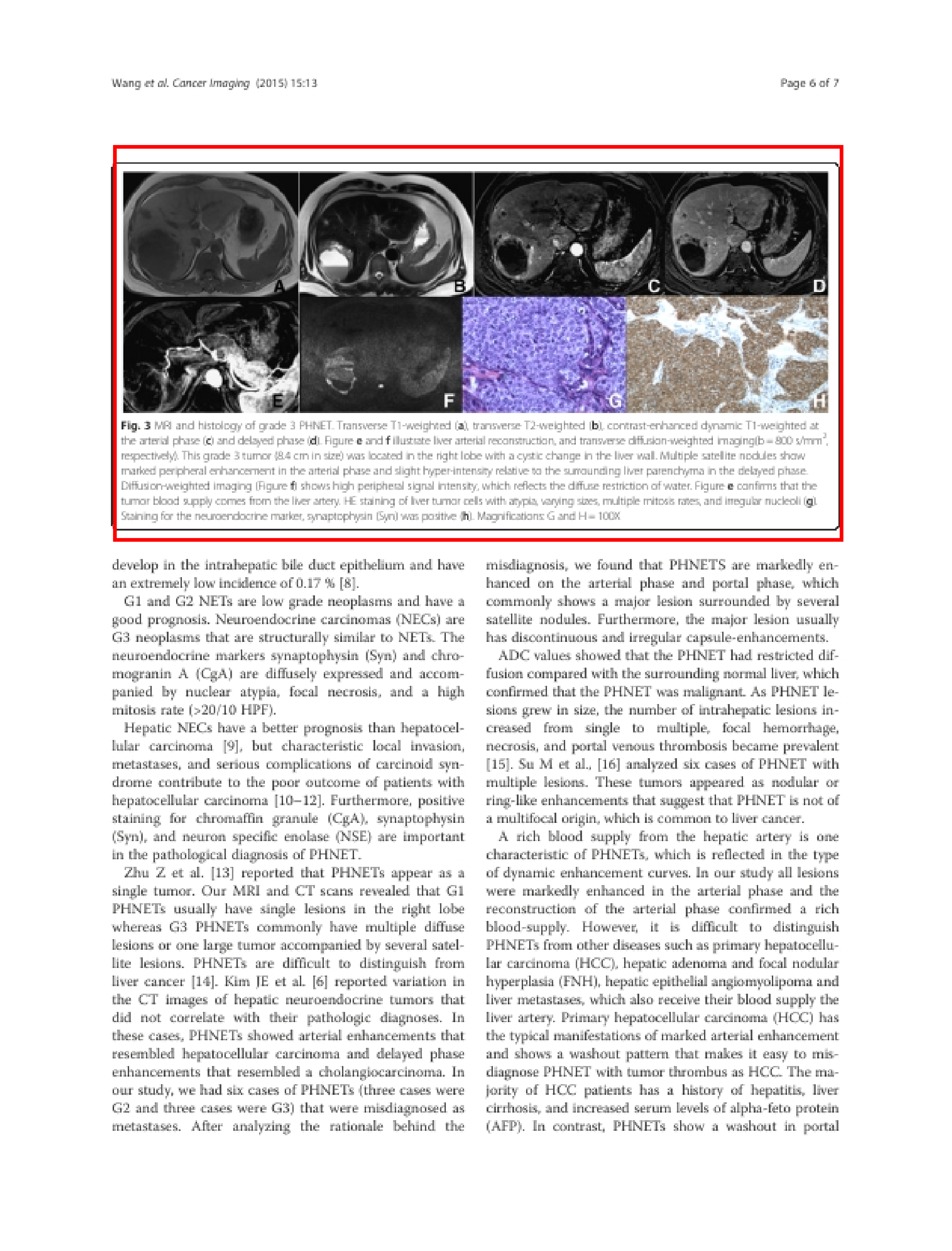}  
    \caption{Pred G}
    \label{p1}
\end{subfigure}
\begin{subfigure}{.47\textwidth}
    \centering
    \includegraphics[width=.98\linewidth]{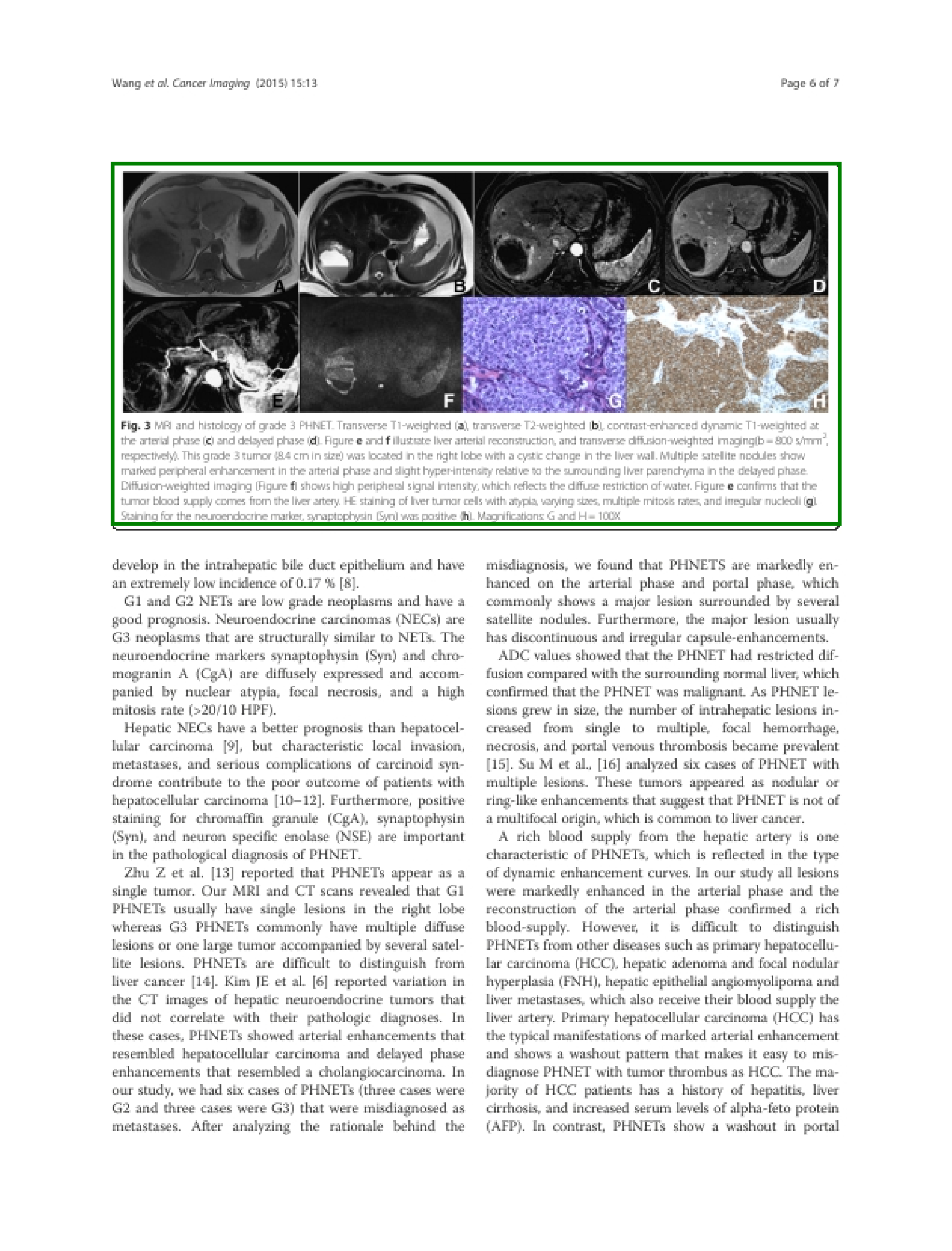}  
    \caption{Target G}
    \label{t1}
\end{subfigure}
\vspace*{2mm}
\begin{subfigure}{.47\textwidth}
    \centering
    \includegraphics[width=.98\linewidth]{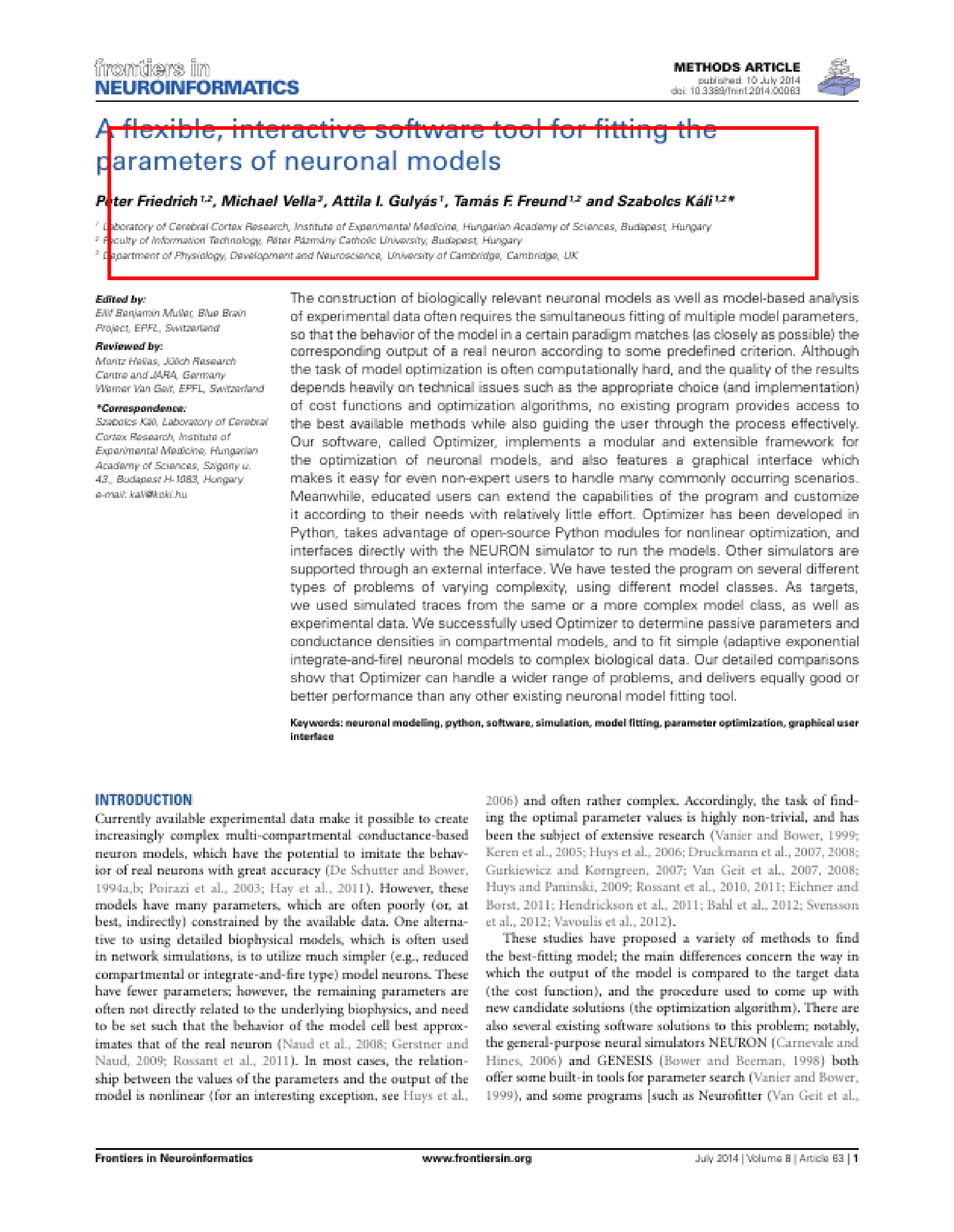}  
    \caption{Pred H}
    \label{q2}
\end{subfigure}
\begin{subfigure}{.47\textwidth}
    \centering
    \includegraphics[width=.98\linewidth]{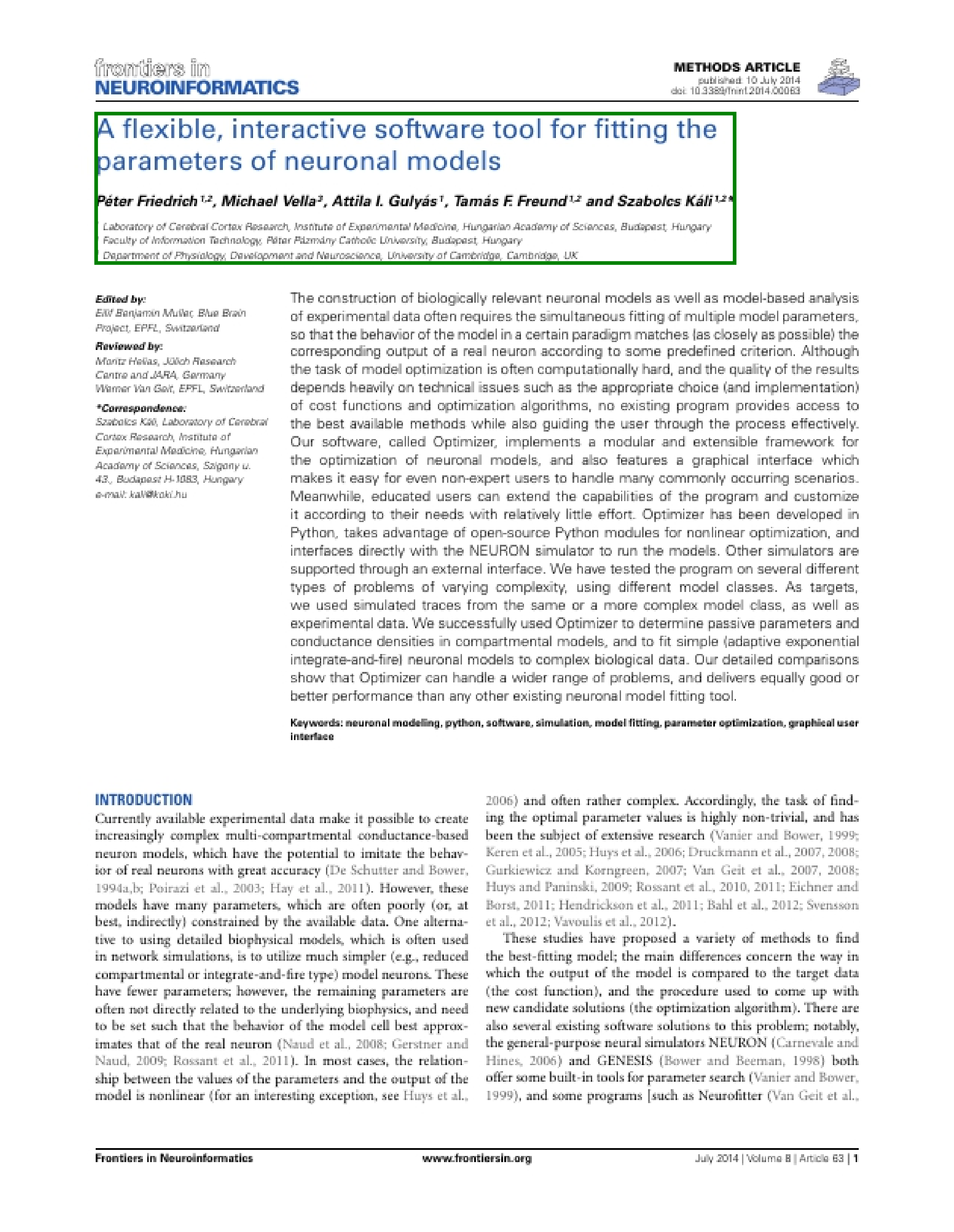}  
    \caption{Target H}
    \label{q2}
\end{subfigure}

\caption{PubLayNet Examples (1)}
\label{pub1}
\end{figure*}

\begin{figure*}
\centering

\begin{subfigure}{.47\textwidth}
    \centering
    \includegraphics[width=.98\linewidth]{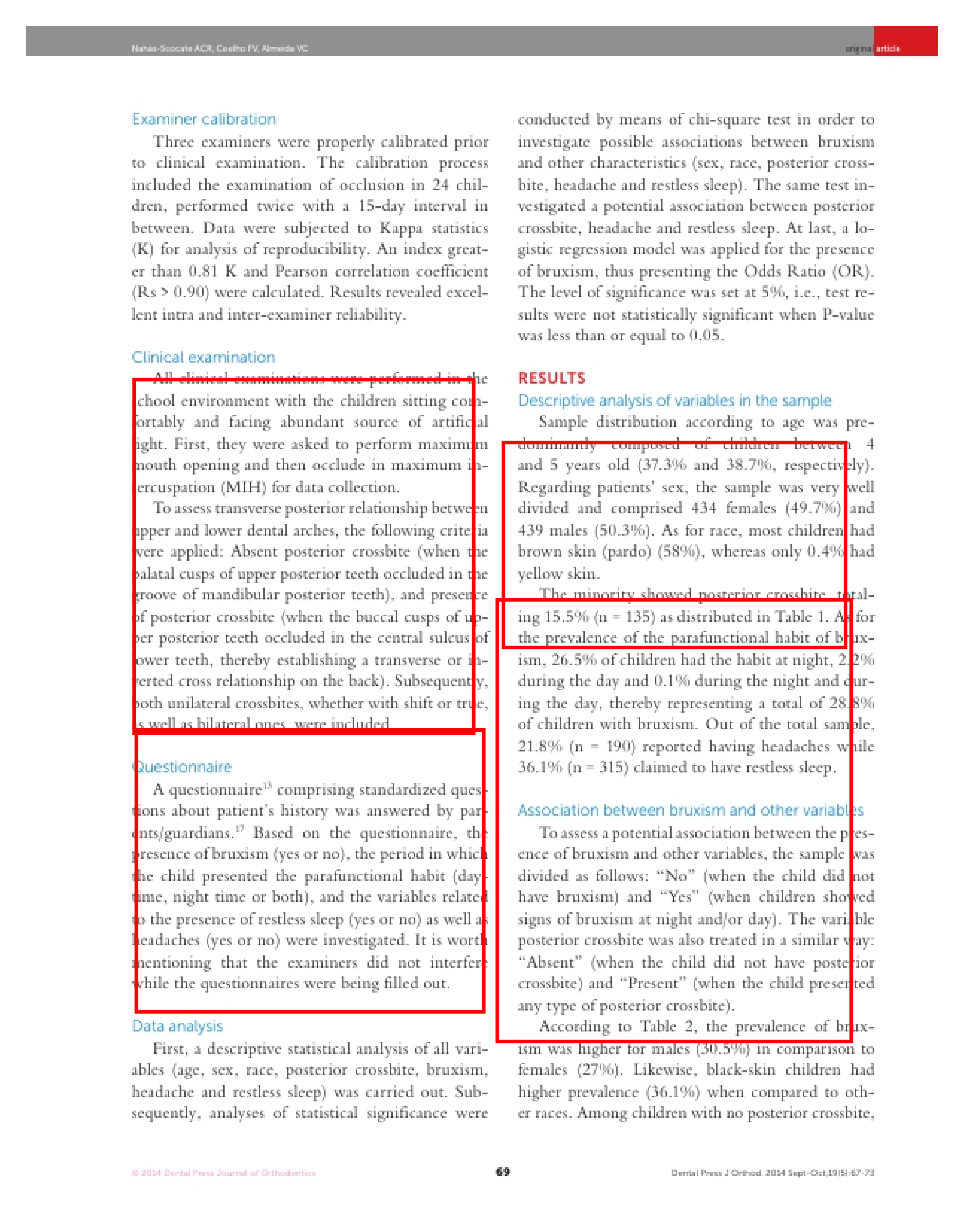}  
    \caption{Pred I}
    \label{p3}
\end{subfigure}
\begin{subfigure}{.47\textwidth}
    \centering
    \includegraphics[width=.98\linewidth]{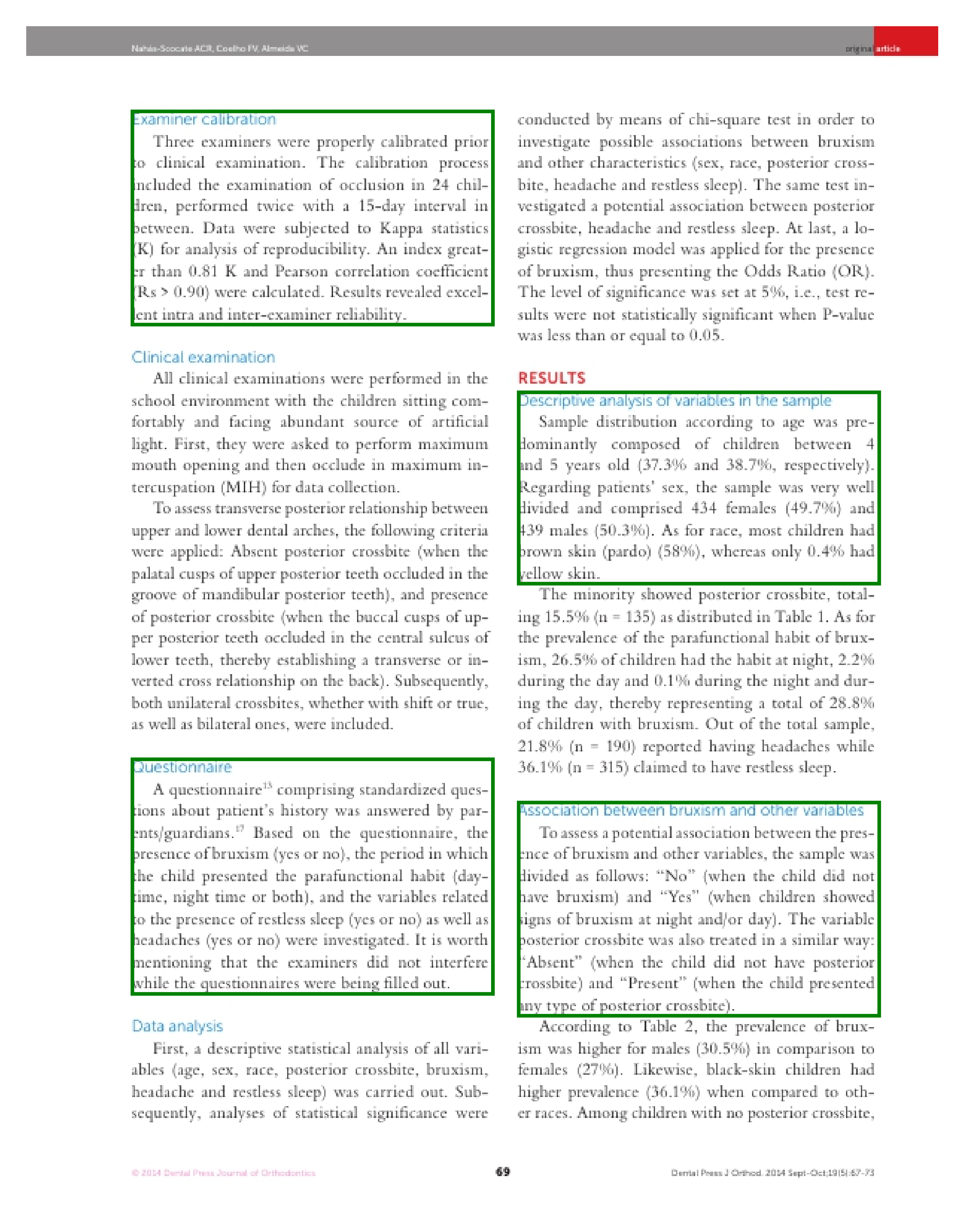}  
    \caption{Target I}
    \label{t3}
\end{subfigure}
\vspace*{2mm}
\begin{subfigure}{.47\textwidth}
    \centering
    \includegraphics[width=.98\linewidth]{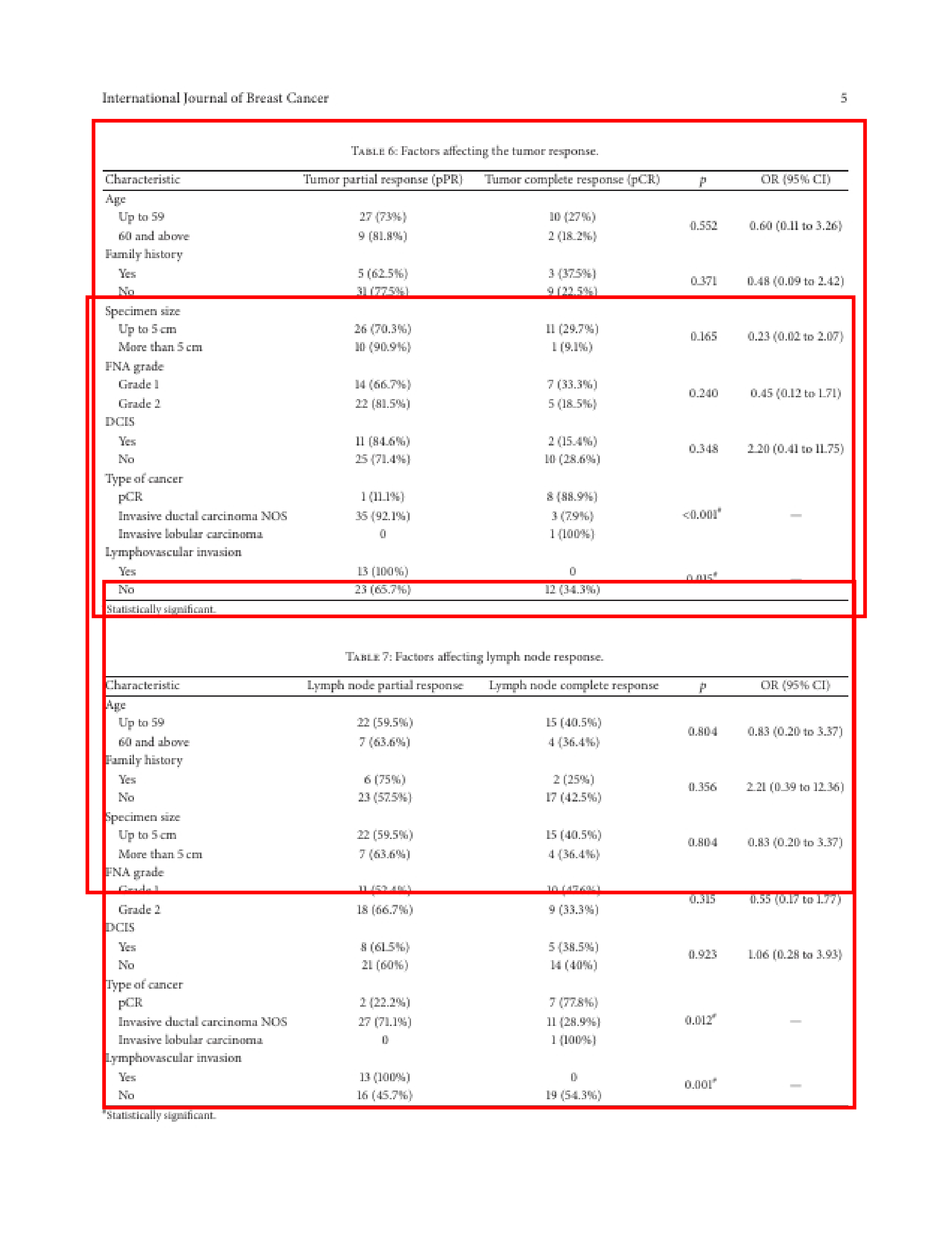}  
    \caption{Pred J}
    \label{q4}
\end{subfigure}
\begin{subfigure}{.47\textwidth}
    \centering
    \includegraphics[width=.98\linewidth]{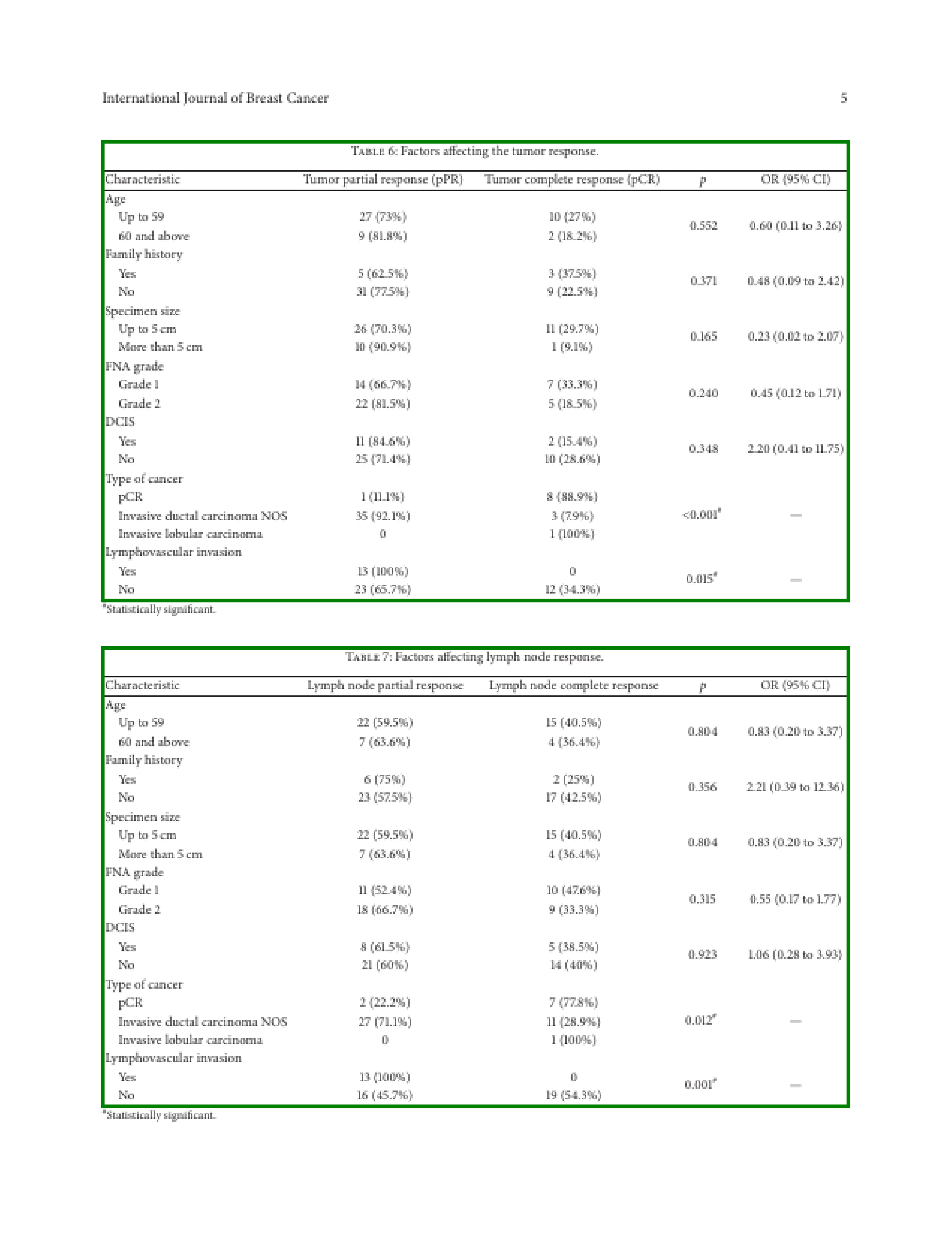}  
    \caption{Target J}
    \label{q4}
\end{subfigure}

\caption{PubLayNet Examples (2)}
\label{pub2}
\end{figure*}

\subsection{Bounding Box Detection}
The $F_{sim}$ is passed through a linear layer followed by a sequence of 4 convolutional layers to produce features which are reshaped to give outputs at 4 different levels as described in the main paper. Then we apply a standard FPN \cite{lin2017feature} to obtain features at a common representation size of $1024$. Finally, we generate proposals most similar to the query through an RPN \cite{girshick2014rich} and subsequently detect bounding boxes using RoI Heads. We choose the default parameters for the RPN and RoI heads (from \cite{torchvision}) --
\begin{itemize}
    \item RPN NMS threshold = $0.7$ 
    \item RPN IOU threshold = $0.7$ (FG), $0.30$ (BG)
    \item RPN Score Threshold = $0$
    \item ROI NMS Threshold = $0.40$
    \item ROI Score Threshold = $0.05$
    \item Detections per image = $200$
    \item ROI IOU threshold = $0.50$ (FG), $0.50$ (BG)
\end{itemize}
where FG, BG are the foreground and background respectively. Note that the bounding box detection for LayoutLMv3 baseline is kept exactly the same as the proposed MONOMER whereas for BHRL, the github implementation is used.

\section{Human Evaluation}
In this section, we delineate the human evaluation conducted on the generated dataset through the proposed technique. A summary of the results have been tabulated in Table \ref{tab:eval}. We create $4$ dataset split containing $40$ samples each and share each split with $3$ human evaluators to report the metrics. The high recall and precision over all the splits indicates that not only does our method generate high quality ground truths ($87.96\%$) but is also able to find most of the target regions ($81.07\%$) in a given document corresponding to a particular query. This saves a considerable amount of human annotation costs while maintaining reliability. Further, we also note that a substantial number of samples ($48.12\%$) over all splits are complex and hard to search for in a document. While this metric is largely subjective, its consistency over multiple splits verifies our claim. Finally, the last row shows that query matching within targets is non-trivial such that only $12.42\%$ cases where snippet and highlighted similar regions are exact matches with the rest of the non-trivial matches containing the same layout but possibly different variation, text, fonts etc which allows the model to learn "advanced one-shot search capabilities".

\section{Additional Qualitative Results}
We add more results produced by the proposed MONOMER in this section. Please refer to ~Fig. \ref{forms1} - ~Fig. \ref{pub2} for the predictions (Page 4 onwards).



{\small
\bibliographystyle{ieee_fullname}
\bibliography{egbib}

\begin{thebibliography}{10}\itemsep=-1pt

\bibitem{oecd2020digital}
OECD 2020.
\newblock Digital transformation in the age of covid-19: Building resilience
  and bridging divides.
\newblock {\em Digital Economy Outlook 2020 Supplement, OECD, Paris,
  www.oecd.org/digital/digital-economy-outlook-covid.pdf}, 2020.

\bibitem{aggarwal-etal-2020-form2seq}
Milan Aggarwal, Hiresh Gupta, Mausoom Sarkar, and Balaji Krishnamurthy.
\newblock {F}orm2{S}eq : A framework for higher-order form structure
  extraction.
\newblock In {\em Proceedings of the 2020 Conference on Empirical Methods in
  Natural Language Processing (EMNLP)}, pages 3830--3840, Online, Nov. 2020.
  Association for Computational Linguistics.

\bibitem{aggarwal2020multi}
Milan Aggarwal, Mausoom Sarkar, Hiresh Gupta, and Balaji Krishnamurthy.
\newblock Multi-modal association based grouping for form structure extraction.
\newblock In {\em Proceedings of the IEEE/CVF Winter Conference on Applications
  of Computer Vision}, pages 2075--2084, 2020.

\bibitem{6831022}
Alireza Alaei and Mathieu Delalandre.
\newblock A complete logo detection/recognition system for document images.
\newblock In {\em 2014 11th IAPR International Workshop on Document Analysis
  Systems}, pages 324--328, 2014.

\bibitem{appalaraju2021docformer}
Srikar Appalaraju, Bhavan Jasani, Bhargava~Urala Kota, Yusheng Xie, and R
  Manmatha.
\newblock Docformer: End-to-end transformer for document understanding.
\newblock In {\em Proceedings of the IEEE/CVF International Conference on
  Computer Vision}, pages 993--1003, 2021.

\bibitem{bao2021beit}
Hangbo Bao, Li Dong, and Furu Wei.
\newblock Beit: Bert pre-training of image transformers.
\newblock {\em arXiv preprint arXiv:2106.08254}, 2021.

\bibitem{Berry2007SurveyOT}
Michael~W. Berry and Mal{\'u} Castellanos.
\newblock Survey of text mining: Clustering, classification, and retrieval.
\newblock 2007.

\bibitem{10.1145/3355610}
Galal~M. Binmakhashen and Sabri~A. Mahmoud.
\newblock Document layout analysis: A comprehensive survey.
\newblock {\em ACM Comput. Surv.}, 52(6), oct 2019.

\bibitem{bowen2009document}
Glenn~A Bowen.
\newblock Document analysis as a qualitative research method.
\newblock {\em Qualitative research journal}, 2009.

\bibitem{10.1145/2071389.2071390}
Claudio Carpineto and Giovanni Romano.
\newblock A survey of automatic query expansion in information retrieval.
\newblock {\em ACM Comput. Surv.}, 44(1), jan 2012.

\bibitem{chen2021adaptive}
Ding-Jie Chen, He-Yen Hsieh, and Tyng-Luh Liu.
\newblock Adaptive image transformer for one-shot object detection.
\newblock In {\em Proceedings of the IEEE/CVF Conference on Computer Vision and
  Pattern Recognition}, pages 12247--12256, 2021.

\bibitem{cheng2019qatm}
Jiaxin Cheng, Yue Wu, Wael AbdAlmageed, and Premkumar Natarajan.
\newblock Qatm: Quality-aware template matching for deep learning.
\newblock In {\em Proceedings of the IEEE/CVF Conference on Computer Vision and
  Pattern Recognition}, pages 11553--11562, 2019.

\bibitem{devlin2018bert}
Jacob Devlin, Ming-Wei Chang, Kenton Lee, and Kristina Toutanova.
\newblock Bert: Pre-training of deep bidirectional transformers for language
  understanding.
\newblock {\em arXiv preprint arXiv:1810.04805}, 2018.

\bibitem{distefano_adobe_2020}
Sebastian Distefano.
\newblock Lessons on active learning in the covid-19 era.
\newblock {\em Adobe Blog,
  https://blog.adobe.com/en/publish/2021/06/22/lessons-on-active-learning-in-the-covid-19-era},
  2021.

\bibitem{searching_pdf_2021}
Adobe Document~Cloud.
\newblock Searching pdfs.
\newblock 2021.
\newblock publisher: Adobe.

\bibitem{document_cloud_how_2022}
Adobe Document~Cloud.
\newblock How to collaborate on a {PDF}., 2022.
\newblock publisher: Adobe.

\bibitem{girshick2014rich}
Ross Girshick, Jeff Donahue, Trevor Darrell, and Jitendra Malik.
\newblock Rich feature hierarchies for accurate object detection and semantic
  segmentation.
\newblock In {\em Proceedings of the IEEE conference on computer vision and
  pattern recognition}, pages 580--587, 2014.

\bibitem{7449303}
M.B. Hisham, Shahrul~Nizam Yaakob, R.A.A Raof, A.B~A. Nazren, and N.M. Wafi.
\newblock Template matching using sum of squared difference and normalized
  cross correlation.
\newblock In {\em 2015 IEEE Student Conference on Research and Development
  (SCOReD)}, pages 100--104, 2015.

\bibitem{hsieh2019one}
Ting-I Hsieh, Yi-Chen Lo, Hwann-Tzong Chen, and Tyng-Luh Liu.
\newblock One-shot object detection with co-attention and co-excitation.
\newblock {\em Advances in neural information processing systems}, 32, 2019.

\bibitem{huang2022layoutlmv3}
Yupan Huang, Tengchao Lv, Lei Cui, Yutong Lu, and Furu Wei.
\newblock Layoutlmv3: Pre-training for document ai with unified text and image
  masking.
\newblock {\em arXiv preprint arXiv:2204.08387}, 2022.

\bibitem{kiran2021offline}
P Kiran, BD Parameshachari, J Yashwanth, and KN Bharath.
\newblock Offline signature recognition using image processing techniques and
  back propagation neuron network system.
\newblock {\em SN Computer Science}, 2(3):1--8, 2021.

\bibitem{li2018high}
Bo Li, Junjie Yan, Wei Wu, Zheng Zhu, and Xiaolin Hu.
\newblock High performance visual tracking with siamese region proposal
  network.
\newblock In {\em Proceedings of the IEEE conference on computer vision and
  pattern recognition}, pages 8971--8980, 2018.

\bibitem{li2022dit}
Junlong Li, Yiheng Xu, Tengchao Lv, Lei Cui, Cha Zhang, and Furu Wei.
\newblock Dit: Self-supervised pre-training for document image transformer,
  2022.

\bibitem{li2021selfdoc}
Peizhao Li, Jiuxiang Gu, Jason Kuen, Vlad~I Morariu, Handong Zhao, Rajiv Jain,
  Varun Manjunatha, and Hongfu Liu.
\newblock Selfdoc: Self-supervised document representation learning.
\newblock In {\em Proceedings of the IEEE/CVF Conference on Computer Vision and
  Pattern Recognition}, pages 5652--5660, 2021.

\bibitem{lin2017feature}
Tsung-Yi Lin, Piotr Doll{\'a}r, Ross Girshick, Kaiming He, Bharath Hariharan,
  and Serge Belongie.
\newblock Feature pyramid networks for object detection.
\newblock In {\em Proceedings of the IEEE conference on computer vision and
  pattern recognition}, pages 2117--2125, 2017.

\bibitem{lin2014microsoft}
Tsung-Yi Lin, Michael Maire, Serge Belongie, James Hays, Pietro Perona, Deva
  Ramanan, Piotr Doll{\'a}r, and C~Lawrence Zitnick.
\newblock Microsoft coco: Common objects in context.
\newblock In {\em European conference on computer vision}, pages 740--755.
  Springer, 2014.

\bibitem{macdonald2001using}
Keith Macdonald.
\newblock Using documents 12.
\newblock {\em Researching social life}, page 194, 2001.

\bibitem{torchvision}
S\'{e}bastien Marcel and Yann Rodriguez.
\newblock Torchvision the machine-vision package of torch.
\newblock In {\em Proceedings of the 18th ACM International Conference on
  Multimedia}, MM '10, page 1485–1488, New York, NY, USA, 2010. Association
  for Computing Machinery.

\bibitem{mathew2022infographicvqa}
Minesh Mathew, Viraj Bagal, Rub{\`e}n Tito, Dimosthenis Karatzas, Ernest
  Valveny, and CV Jawahar.
\newblock Infographicvqa.
\newblock In {\em Proceedings of the IEEE/CVF Winter Conference on Applications
  of Computer Vision}, pages 1697--1706, 2022.

\bibitem{melekhov2016siamese}
Iaroslav Melekhov, Juho Kannala, and Esa Rahtu.
\newblock Siamese network features for image matching.
\newblock In {\em 2016 23rd international conference on pattern recognition
  (ICPR)}, pages 378--383. IEEE, 2016.

\bibitem{10.5555/1822502}
Frederic~P. Miller, Agnes~F. Vandome, and John McBrewster.
\newblock {\em Levenshtein Distance: Information Theory, Computer Science,
  String (Computer Science), String Metric, Damerau?Levenshtein Distance, Spell
  Checker, Hamming Distance}.
\newblock Alpha Press, 2009.

\bibitem{modi_adobe_2020}
Abhigyan Modi.
\newblock India’s move to digital documents in light of covid-19.
\newblock {\em Adobe Blog,
  https://blog.adobe.com/en/publish/2020/05/19/indias-move-to-digital-documents-in-light-of-covid-19},
  2020.

\bibitem{neubeck2006efficient}
Alexander Neubeck and Luc Van~Gool.
\newblock Efficient non-maximum suppression.
\newblock In {\em 18th International Conference on Pattern Recognition
  (ICPR'06)}, volume~3, pages 850--855. IEEE, 2006.

\bibitem{oyallon2015analysis}
Edouard Oyallon and Julien Rabin.
\newblock An analysis of the surf method.
\newblock {\em Image Processing On Line}, 5:176--218, 2015.

\bibitem{redmon2016you}
Joseph Redmon, Santosh Divvala, Ross Girshick, and Ali Farhadi.
\newblock You only look once: Unified, real-time object detection.
\newblock In {\em Proceedings of the IEEE conference on computer vision and
  pattern recognition}, pages 779--788, 2016.

\bibitem{reed2022generalist}
Scott Reed, Konrad Zolna, Emilio Parisotto, Sergio~Gomez Colmenarejo, Alexander
  Novikov, Gabriel Barth-Maron, Mai Gimenez, Yury Sulsky, Jackie Kay,
  Jost~Tobias Springenberg, et~al.
\newblock A generalist agent.
\newblock {\em arXiv preprint arXiv:2205.06175}, 2022.

\bibitem{ren2015faster}
Shaoqing Ren, Kaiming He, Ross Girshick, and Jian Sun.
\newblock Faster r-cnn: Towards real-time object detection with region proposal
  networks.
\newblock {\em Advances in neural information processing systems}, 28, 2015.

\bibitem{ruder2016overview}
Sebastian Ruder.
\newblock An overview of gradient descent optimization algorithms.
\newblock {\em arXiv preprint arXiv:1609.04747}, 2016.

\bibitem{ruff2018deep}
Lukas Ruff, Robert Vandermeulen, Nico Goernitz, Lucas Deecke, Shoaib~Ahmed
  Siddiqui, Alexander Binder, Emmanuel M{\"u}ller, and Marius Kloft.
\newblock Deep one-class classification.
\newblock In {\em International conference on machine learning}, pages
  4393--4402. PMLR, 2018.

\bibitem{sarkar2020document}
Mausoom Sarkar, Milan Aggarwal, Arneh Jain, Hiresh Gupta, and Balaji
  Krishnamurthy.
\newblock Document structure extraction using prior based high resolution
  hierarchical semantic segmentation.
\newblock In {\em European Conference on Computer Vision}, pages 649--666.
  Springer, 2020.

\bibitem{song2020mpnet}
Kaitao Song, Xu Tan, Tao Qin, Jianfeng Lu, and Tie-Yan Liu.
\newblock Mpnet: Masked and permuted pre-training for language understanding.
\newblock {\em Advances in Neural Information Processing Systems},
  33:16857--16867, 2020.

\bibitem{tito2021icdar}
Rub{\`e}n Tito, Minesh Mathew, CV Jawahar, Ernest Valveny, and Dimosthenis
  Karatzas.
\newblock Icdar 2021 competition on document visual question answering.
\newblock In {\em International Conference on Document Analysis and
  Recognition}, pages 635--649. Springer, 2021.

\bibitem{vaswani2017attention}
Ashish Vaswani, Noam Shazeer, Niki Parmar, Jakob Uszkoreit, Llion Jones,
  Aidan~N Gomez, {\L}ukasz Kaiser, and Illia Polosukhin.
\newblock Attention is all you need.
\newblock {\em Advances in neural information processing systems}, 30, 2017.

\bibitem{wu2013comparative}
Jian Wu, Zhiming Cui, Victor~S Sheng, Pengpeng Zhao, Dongliang Su, and
  Shengrong Gong.
\newblock A comparative study of sift and its variants.
\newblock {\em Measurement science review}, 13(3), 2013.

\bibitem{xu2020layoutlm}
Yiheng Xu, Minghao Li, Lei Cui, Shaohan Huang, Furu Wei, and Ming Zhou.
\newblock Layoutlm: Pre-training of text and layout for document image
  understanding.
\newblock In {\em Proceedings of the 26th ACM SIGKDD International Conference
  on Knowledge Discovery \& Data Mining}, pages 1192--1200, 2020.

\bibitem{yang2022balanced}
Hanqing Yang, Sijia Cai, Hualian Sheng, Bing Deng, Jianqiang Huang, Xian-Sheng
  Hua, Yong Tang, and Yu Zhang.
\newblock Balanced and hierarchical relation learning for one-shot object
  detection.
\newblock In {\em Proceedings of the IEEE/CVF Conference on Computer Vision and
  Pattern Recognition}, pages 7591--7600, 2022.

\bibitem{yang2017learning}
Xiao Yang, Ersin Yumer, Paul Asente, Mike Kraley, Daniel Kifer, and C
  Lee~Giles.
\newblock Learning to extract semantic structure from documents using
  multimodal fully convolutional neural networks.
\newblock In {\em Proceedings of the IEEE Conference on Computer Vision and
  Pattern Recognition}, pages 5315--5324, 2017.

\bibitem{yoo2009fast}
Jae-Chern Yoo and Tae~Hee Han.
\newblock Fast normalized cross-correlation.
\newblock {\em Circuits, systems and signal processing}, 28(6):819--843, 2009.

\bibitem{zaidi2022survey}
Syed Sahil~Abbas Zaidi, Mohammad~Samar Ansari, Asra Aslam, Nadia Kanwal,
  Mamoona Asghar, and Brian Lee.
\newblock A survey of modern deep learning based object detection models.
\newblock {\em Digital Signal Processing}, page 103514, 2022.

\bibitem{zhong2019publaynet}
Xu Zhong, Jianbin Tang, and Antonio~Jimeno Yepes.
\newblock Publaynet: largest dataset ever for document layout analysis.
\newblock In {\em 2019 International Conference on Document Analysis and
  Recognition (ICDAR)}, pages 1015--1022. IEEE, Sep. 2019.

\end{thebibliography}
}

\end{document}